\pdfoutput=1

\documentclass[10pt,twocolumn,letterpaper]{article}

\usepackage[pagenumbers]{wacv} 

\usepackage{makecell}   
\usepackage{float}      
\usepackage{placeins}   
\usepackage{colortbl}   
\usepackage[ruled]{algorithm2e}  
\usepackage{cuted}
\usepackage{balance}    

\SetAlFnt{\small}
\SetAlCapFnt{\small}
\SetAlCapNameFnt{\small}
\SetAlCapHSkip{0pt}

\definecolor{wacvblue}{rgb}{0.21,0.49,0.74}
\usepackage[breaklinks,colorlinks,allcolors=wacvblue]{hyperref}

\def\wacvPaperID{880} 
\def\confName{WACV}
\def\confYear{2027}

\title{\textcolor{Purple}{RADI3N}: Geomet\textcolor{Purple}{R}ic Im\textcolor{Purple}{A}ge E\textcolor{Purple}{D}iting via Parametr\textcolor{Purple}{I}c \textcolor{Purple}{3}D Co\textcolor{Purple}{N}trol}

\author{
Abdul Wasi$^{1}$ \qquad Akhil Gorugantu$^{1}$ \qquad Mahesh Bhosale$^{1}$ \qquad Anjan Dutta$^{2}$ \\
Junsong Yuan$^{1}$ \qquad Sauradip Nag$^{2,3}$\thanks{corresponding author}
\qquad David Doermann$^{1}$ \\[6pt]
{ $^{1}$University at Buffalo \qquad $^{2}$University of Surrey \qquad $^{3}$Simon Fraser University} \\[2pt]
}

\begin{document}
\maketitle

\begin{strip}
\centering
\includegraphics[width=\textwidth]{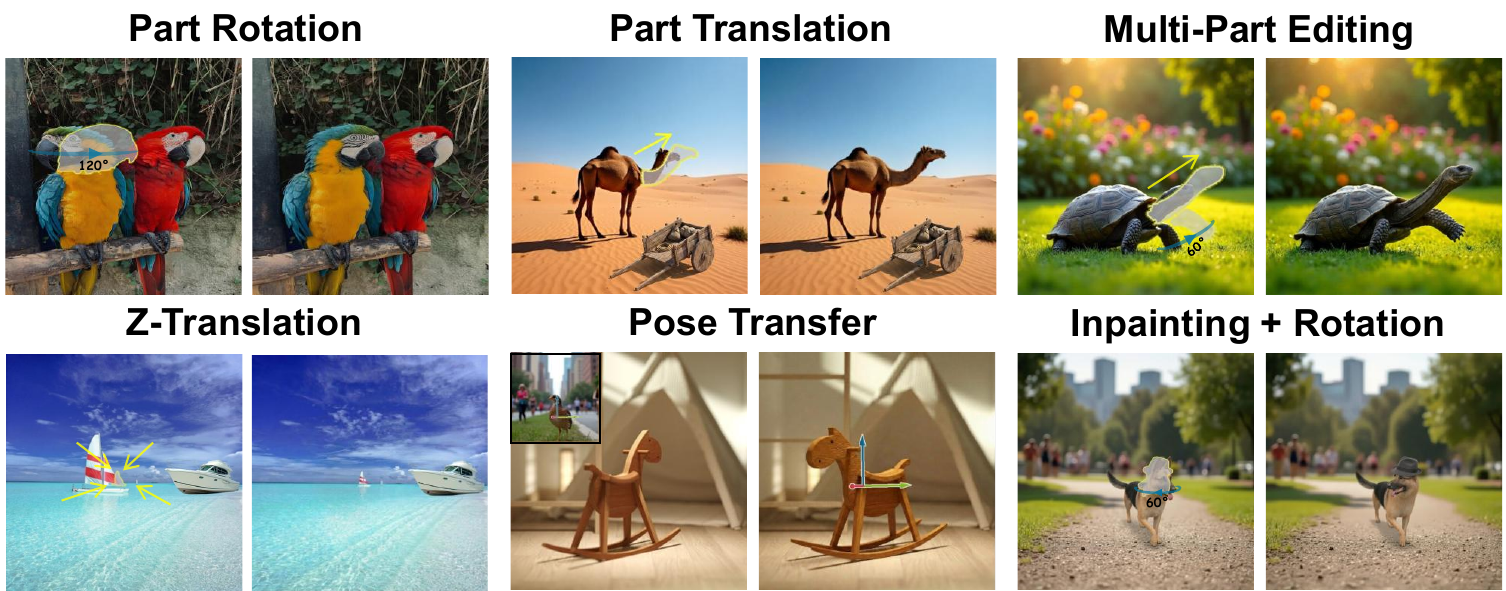}

\vspace{-2mm}

\captionof{figure}{\textbf{RADI3N enables parametric Geometric Image Editing.}
Given an image, our method supports a diverse range of geometric editing
operations, including part rotation, part translation, multi-part editing,
scaling, pose transfer, and inpainting with rotation, without any
task-specific training or fine-tuning.}
\label{fig:teaser}
\end{strip}

\begin{abstract}
Recent diffusion-based methods have substantially advanced image editing, yet parametric 3D control at the part level remains challenging. Existing approaches either operate on whole objects with manual rigging, or revert to 2D handles at the part level. They also struggle to model appearance and texture under large transformations, leading to unrealistic outputs in geometry-intensive edits. To address these limitations, we introduce \textbf{RADI3N}, a framework for semantically-grounded part-aware object manipulation with parametric 3D control. Our CS-ARAP deformation module suppresses rotation-field discontinuities at part boundaries via neighborhood covariance smoothing, producing structurally consistent deformations under complex part-level edits. We further introduce an inversion-free selective texture transfer mechanism that suppresses high-frequency positional biases and scales up low-frequency semantic cues within cross-image attention. This eliminates strict spatial bias and supports semantic matching, allowing source appearance to follow the edited part onto its new geometry. Experiments demonstrate that RADI3N achieves state-of-the-art performance in geometric fidelity, viewpoint consistency, and controllability. Code: \url{https://github.com/wasilone11/Radi3n}
\end{abstract}
\vspace{-20pt}
\section{Introduction}
\label{sec:introduction}

Image generation models now produce photorealistic, high-fidelity images~\cite{hurst2024gpt,team2025gemma,midjourney2025,labs2025flux,adobefirefly,balaji2022eDiff-I,rombach2021highresolution}, making controllable image editing important for film post-production, augmented reality, and creative design, domains that demand parametric control over object attributes such as pose, shape, position, and layout. 
However, manipulating object geometry in natural images remains challenging, as such edits can introduce shape distortions, inconsistent shadows, and mismatches with the surrounding scene. We therefore focus on geometry-consistent 3D editing of natural images, aiming to bridge the flexibility of 2D image editing with the structural control of 3D workflows.

Existing methods rely on text prompts~\cite{meng2022sdedit} or 2D
spatial
controls~\cite{patashnik2023localizing,mirzaei2023watchyoursteps,
hertz2023delta,Nam_2024_CVPR} with limited geometric reasoning,
struggling with coherent 3D transformations under large pose changes.
Recent geometric editing
methods~\cite{yenphraphai2024image,michel2023object,zhu2025training,
koo2025flowdrag,cheng20253d} differ along two axes
(Table~\ref{tab:capability_matrix}): control granularity (whole-object
vs.\ part-level) and control signal (2D displacement vs.\ 3D
parametric). Drag-based
approaches~\cite{shi2024dragdiffusion,koo2025flowdrag,
zhang2024gooddrag} propagate 2D handles but lack semantic part
grounding and fail under large rotations. Lifting-based
methods~\cite{yenphraphai2024image,zhu2025training,michel2023object}
enable parametric 3D control only at whole-object granularity: Image
Sculpting~\cite{yenphraphai2024image} demands manual rigging, while
FreeFine~\cite{zhu2025training} reverts to 2D drag handles at the part
level, forfeiting 3D consistency. Moreover, re-projecting edited mesh
renders introduces foreground-background blending artifacts. These
gaps motivate a pipeline coupling semantic part understanding,
parametric 3D control, and photorealistic re-integration.

To address these requirements, we introduce \textbf{RADI3N}, a
framework for semantically-grounded parametric 3D part-level editing
from a single image. The input to RADI3N is a natural-language description of the target part
and a 6-DOF transformation, automatically lifting the object into a
part-aware 3D representation. Applying the transformation, however, is
non-trivial: standard ARAP induces rotation spikes at part
boundaries~\cite{levi2014smooth} that mesh refinement cannot recover
from. Our CS-ARAP (Covariance-Smoothed ARAP) suppresses them by
averaging neighborhood covariances before SO(3) projection, after which
we refine global geometry, including newly exposed surfaces, by
distilling a pre-trained diffusion prior~\cite{poole2022dreamfusion}.
Accurate geometry alone is not enough: the lifted
representation carries only a coarse texture without material or
illumination, so blending the edited object back to image space re-renders it with flattened materials
and compositing seams~\cite{yenphraphai2024image}. We therefore perform
re-integration as \emph{appearance correction}, transferring the
appearance of the same object instance from the source image onto its
edited counterpart. Attention sharing~\cite{hertz2024style} is a
natural mechanism for such cross-image restoration, but in DiTs the
positional bias of rotary embeddings~\cite{su2024roformer} forces target tokens to copy from
spatially aligned source locations~\cite{zhang2026alignedgen}. 
In our setting, appearance is therefore erroneously copied from the part's pre-edit location. To resolve this, our inversion-free modulated texture transfer applies a mask over the source keys that suppresses their high-frequency positional components and scales up the low-frequency ones. This eliminates strict spatial bias while strengthening the low-frequency components of RoPE that support semantic matching, allowing appearance to seamlessly follow the part onto its new geometry. The restored appearance thus binds seamlessly to the edited geometry, yielding photorealistic edits with consistent texture and illumination in the original scene.

\begin{table}[t]
\centering
\caption{\textbf{Geometric editing capability matrix.} Existing methods cover 2D editing and whole-object 3D parametric control. The combination of 3D parametric control with part-level granularity is uniquely addressed by RADI3N.}
\label{tab:capability_matrix}
\footnotesize
\setlength{\tabcolsep}{4pt}
\renewcommand{\arraystretch}{1.3}
\resizebox{\linewidth}{!}{%
\begin{tabular}{@{}l >{\raggedright\arraybackslash}p{3.2cm} >{\raggedright\arraybackslash}p{3.2cm}@{}}
\toprule
 & \textbf{Whole-object} & \textbf{Part-level} \\
\midrule
\textbf{2D control} &
FreeFine~\cite{zhu2025training}, FlowDrag~\cite{koo2025flowdrag}, GoodDrag~\cite{zhang2024gooddrag} &
FreeFine~\cite{zhu2025training}, Qwen-Img-Edit~\cite{qwen_image_edit}, GoodDrag~\cite{zhang2024gooddrag} \\
\midrule
\textbf{\makecell[l]{Parametric 3D \\ control}} &
FreeFine~\cite{zhu2025training}, Object3DiT~\cite{michel2023object}, 3D-Fixup~\cite{cheng20253d}, Image Sculpting~\cite{yenphraphai2024image}, \textbf{RADI3N} &
\cellcolor[gray]{0.9}\textbf{RADI3N (automatic)}, Image Sculpting~\cite{yenphraphai2024image} (manual) \\
\bottomrule
\end{tabular}%
}
\end{table}

Our contributions are:
\noindent\textbf{(1)} We introduce RADI3N, a framework for semantically-grounded, parametric 3D part-level image editing from a single image. Users specify edits via natural language and exact 6-DOF transformations, enabling precise part-level geometric control without manual rigging.
\noindent\textbf{(2)} We propose CS-ARAP, a deformation energy that suppresses
rotation-field discontinuities at part boundaries via neighborhood
covariance averaging before SO(3) projection, and refine the deformed
mesh by distilling a pre-trained diffusion prior, enabling
artifact-free deformations under complex 3D part edits.
\noindent\textbf{(3)} We reformulate photorealistic blending as \emph{appearance correction} using an inversion-free modulated texture transfer. By suppressing high-frequency positional biases in cross-image attention, we seamlessly restore instance-specific texture and illumination onto the edited geometry, eliminating common reprojection artifacts.
\noindent\textbf{(4)} Extensive evaluation on DragBench~\cite{shi2024dragdiffusion} and GeoBench~\cite{zhu2025training} demonstrates state-of-the-art geometric fidelity, angular accuracy, and photorealism, supported by user studies and workflow demonstrations.

\vspace{-5pt}
\section{Related Work}
\label{sec:related_works}

\subsubsection*{\textbf{Image Editing with Diffusion Models}}
Diffusion models are state-of-the-art for high-fidelity visual generation~\cite{balaji2022eDiff-I,imagen2022saharia,rombach2021highresolution,ramesh2022dalle}, iteratively denoising Gaussian noise into images~\cite{sohl2015diffusion,ho2020ddpm}. Early editing methods added noise to an image and denoised it via text prompts~\cite{meng2022sdedit}, but struggled to balance content preservation with prompt adherence. Later work addressed this via editing masks~\cite{avrahami2020blendeddiffusion,safaee2023clic,patashnik2023localizing,mirzaei2023watchyoursteps,hertz2023delta,Nam_2024_CVPR} and adapter-based frameworks like ControlNet~\cite{zhang2023adding} and T2I-Adapter~\cite{mou2024t2i} for spatial control, as well as instruction-tuned models~\cite{brooks2022instructpix2pix}, per-image optimization~\cite{kawar2023imagic}, and pose/perspective extensions~\cite{cao2023masactrl,koo2024flexiedit}. More recently, the shift from UNet to Diffusion Transformer (DiT) backbones~\cite{peebles2023dit} has enabled more robust editing methods~\cite{zhang2025icedit,song2025insertanything}. Given the cost of fine-tuning, the field has shifted toward training-free techniques~\cite{han2024proxedit,zhu2025kv,avrahami2025stable}. 

\begin{figure*}[t]
    \centering
    \includegraphics[width=0.9\linewidth]{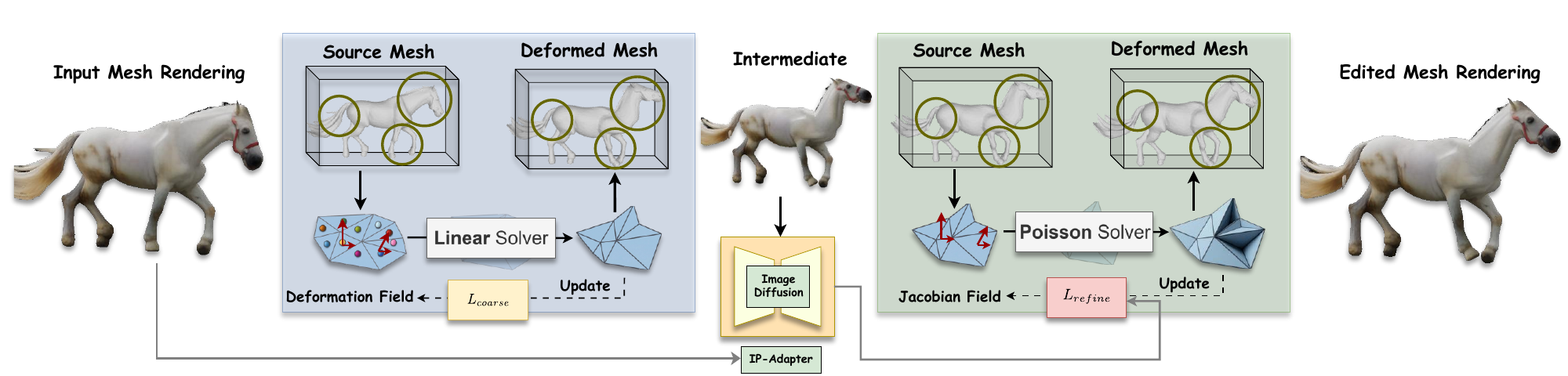}
    \caption{\textbf{Overview of the geometric editing pipeline.} \textbf{Stage 1 (Coarse 3D Editing):} The input image is lifted to a 3D mesh and deformed using CS-ARAP with covariance smoothing and a bi-Laplacian solve ($L_{\text{coarse}}$) to obtain a smooth mesh. Intermediate renderings and source views condition the diffusion model via IP-Adapter for identity preservation. \textbf{Stage 2 (Global 3D Refinement):} The Stage 1 Jacobian field is refined using $L_{\text{refine}}$, combining handle constraints with SDS gradients from the diffusion model, producing a semantically plausible mesh for final rendering.}
    \label{fig:placeholder}
\end{figure*}

\subsubsection*{\textbf{3D-Aware Generative Image Editing}}
The field has progressed toward 3D-aware representations~\cite{kerbl20233d,
liu2023zero,mildenhall2021nerf,voleti2024sv3d} that enable geometric
modifications by rebuilding scenes in 3D space~\cite{yenphraphai2024image}, though these typically require multi-view consistency or costly per-scene
optimization. Recent single-view 3D reconstruction
methods~\cite{liu2023zero,voleti2024sv3d,zhao2025hunyuan3d,sam3d}
have reduced this burden by enabling lift-edit-reproject pipelines from a
single image. In parallel, latent-space manipulation approaches guide
structural edits via drag-based trajectories~\cite{cui2024stabledrag,
li2024dragapart,liu2024drag,pan2023drag,shi2024dragdiffusion,shin2024instantdrag},
segmentation masks~\cite{epstein2023diffusion,pandey2024diffusion,
sajnani2025geodiffuser}, or geometric signals such as optical flow and
scene layouts~\cite{geng2024motion,jia2024designedit,lu2024regiondrag,
mou2023dragondiffusion,mu2024editable}. However, these methods
struggle with large viewpoint shifts and lack precise parametric control.
Our method addresses this by combining single-image 3D
lifting~\cite{voleti2024sv3d,liu2023zero,sam3d,zhao2025hunyuan3d}
with VLM-based instruction parsing~\cite{Qwen3VL} to translate text
prompts into accurate, parametric 3D editing signals.

\newcommand{\bwd}{\text{bw}}
\newcommand{\fwd}{\text{fw}}
\newcommand{\aln}{\text{aln}}
\vspace{-5pt}
\section{Preliminaries}
\label{sec:preliminary}

\subsubsection*{\textbf{Score Distillation Sampling (SDS)}}
SDS~\cite{poole2022dreamfusion} optimizes a 3D representation by distilling
a pre-trained 2D diffusion model $\epsilon_\phi$. Given a rendered image
$I$, the gradient with respect to parameters $\theta$ is:
\begin{equation}
    \nabla_\theta \mathcal{L}_{\text{SDS}} = \mathbb{E}_{t, \epsilon}
    \left[ w(t) \left( \epsilon_\phi(I_t; y, t) - \epsilon \right)
    \frac{\partial I}{\partial \theta} \right],
\end{equation}
where $I_t$ is the noised rendering, $y$ is a text prompt, $w(t)$ a
weighting function, $\theta$ the parameters of the 3D representation being
optimized, and $\epsilon$ the added noise. This pushes the
geometry toward configurations whose renderings lie on the natural image
manifold of the diffusion prior.

\subsubsection*{\textbf{As-Rigid-As-Possible (ARAP) Deformation}}
Given a triangle mesh $M = (V, F)$ with vertices $v_i \in \mathbb{R}^3$,
source vertex positions $v_i, v_j \in V$, the deformed vertices
$\hat{v}_i, \hat{v}_j \in \hat{V}$, ARAP~\cite{sorkine2007rigid} computes
a deformed configuration $\hat{V}$ by minimizing deviation from rigid
local transformations:
\begin{equation}
    E_{\text{ARAP}} = \sum_{i \in V} \sum_{j \in \mathcal{N}_i} w_{ij}
    \| (\hat{v}_i - \hat{v}_j) - R_i (v_i - v_j) \|^2,
\end{equation}
where $R_i \in SO(3)$ is the optimal local rotation, $w_{ij}$ are
cotangent weights, and $\mathcal{N}_i$ is the 1-ring neighborhood.
Optimization alternates between a \textit{local step} computing each
$R_i$ via SVD of the covariance matrix $S_i = \sum_{j \in \mathcal{N}_i}
w_{ij}(v_i - v_j)(\hat{v}_i - \hat{v}_j)^T$, and a \textit{global step}
solving a sparse linear system under fixed boundary constraints.
SR-ARAP~\cite{levi2014smooth} augments this with a rotation consistency
term $\alpha \sum_{i} \sum_{j \in \mathcal{N}_i} \|R_i - R_j\|^2$ that
penalizes discontinuities in the rotation field.

\subsubsection*{\textbf{Discrete Differential Operators}}
The \textit{cotangent Laplacian} $L$ encodes pairwise vertex relationships
via cotangent weights, and the diagonal \textit{mass matrix} $M$ stores
per-vertex area weights. Their composition yields the \textit{bi-Laplacian}
$LM^{-1}L$, which penalizes variation in the position field and enforces
$C^2$ continuity.

\subsubsection*{\textbf{Jacobian Fields and Poisson Reconstruction}}
Deformation is parameterized via a
per-face Jacobian field $\mathcal{J} = \{J_f \in \mathbb{R}^{3\times3}
\}_{f \in F}$, initialized as $J_f = \nabla_f V_0$ from the source mesh
and treated as a trainable parameter. Deformed positions are recovered by:
\begin{equation}
    V^* = \arg\min_V \|LV - \nabla^T \mathcal{A} \mathcal{J}\|^2
    + \lambda \|K_a V - T_a\|^2,
\end{equation}
where $\nabla$ is the stacked per-face gradient operator, $\mathcal{A}$
the diagonal per-face area matrix, $K_a$ the anchor selection matrix,
and $T_a$ the anchor targets. The differentiable solve allows upstream
gradients to propagate directly to $\mathcal{J}$.

\section{Method}
\label{sec:method}
We introduce RADI3N, a diffusion-based framework for part-based parametric image editing with precise 3D geometric constraints and photorealistic consistency. Given an input RGB image $I$, and a user-specified object prompt $P_{obj}$ with a 6-DOF target pose $H_{pose} \in \mathbb{R}^{4 \times 4}$, our Object Abstraction module (Sec.\ref{sec:correspondence}) first lifts the target object to 3D and decomposes it into editable parts. The 3D Editing module (Sec. \ref{sec:vpgpe}) then applies visually plausible geometric transformations to these segmented 3D components, yielding an edited 3D representation. Finally, the Appearance Blending module (Sec. \ref{sec:tex_trans_sec}) renders the edited 3D representation back into the 2D image domain to produce visually realistic editing results.

\vspace{-5pt}
\subsection{Geometric Object Abstraction}
\label{sec:correspondence}
Given an input image $I$ and object prompt $P_{obj}$, we perform \textit{geometric object abstraction} to obtain an editable 3D representation. This stage localizes the object, lifts it to 3D, and decomposes it into meaningful parts. 

\noindent \textbf{Object Selection.}
Given a user editing prompt $P_{obj}$ (e.g., ``Rotate the head of the parrot on the left'' in Fig. 1), we employ a VLM (Vision-Language Model)~\cite{Qwen3VL}) to simultaneously isolate the target object from the input scene and extract the set of semantic parts $P_{parts}$ (e.g., \{``head'', ``ears'', ``neck''\}) relevant to the requested manipulation. To accommodate 6-DoF spatial editing commands, we elevate the isolated 2D object into a 3D representation, denoted as $\mathcal{M}_{obj}$, utilizing an Image-to-3D model \cite{sam3d}.
See supplementary for master prompts.

\noindent \textbf{Part-Aware 3D Editing Handles.}
To enable localized, part-aware 3D editing, we first segment a given 3D mesh $\mathcal{M}_{obj}$ into fine-grained semantic regions based on predefined object parts $P_{parts}$, which facilitates subsequent handle-based 3D deformation. Specifically, we render $\mathcal{M}_{obj}$ from multiple camera viewpoints to acquire an image set $I_{mv} = \{I_{view1}, \ldots, I_{viewN}\}$. Each view is processed independently using \cite{sam3} to generate multi-view 2D part segmentations $\mathbf{I^{seg}_{mv}}$. Because these predictions are inferred independently, they inevitably suffer from multi-view inconsistencies, particularly along part boundaries.
To resolve these discrepancies and lift the 2D segmentations $\mathbf{I^{seg}_{mv}}$ to 3D, we leverage the mesh's \textit{UV}-parameterization. By projecting the per-view segmentations into a shared \textit{UV}-space, we obtain a set of partial \textit{UV}-atlases $\mathbf{A}=\{a_1,\ldots,a_m\}$, where $a_i=\mathbf{UV}(I^{seg}_{mv}[i])$. Since each atlas captures only the visible regions from its respective viewpoint, we fuse them into a single, globally consistent segmentation mask $\mathcal{A}$ via majority voting:
\begin{equation}
\label{eq:uv}
\mathcal{A} = \phi(\mathbf{A}) = \phi(\mathbf{UV}(\mathbf{I^{seg}_{mv}})),
\end{equation}
where $\phi(\cdot)=\texttt{mode()}$. This \textit{UV}-based consensus mitigates boundary artifacts, yielding a view-independent, geometry-agnostic 3D part segmentation suitable for precise handle extraction.

To enable part-aware 3D deformation, we extract sparse geometric handles from segmented parts. Deforming all vertices is inefficient and distortion-prone, so we adopt a keypoint-based paradigm on a compact control-point set. For each view in $I^{seg}_{mv}$, we detect 2D keypoints $I^{key}_{mv}$ within object parts via fast cone singularity~\cite{li2023efficient}. These are lifted to 3D vertices $V_{handle}$ using the UV correspondence in Eq.~\ref{eq:uv}, serving as sparse constraints enabling precise part-aware edits while preserving global structure.

\subsection{Visually Plausible Geometric Part Editing}
\label{sec:vpgpe}
While ARAP~\cite{sorkine2007rigid} preserves local surface details, it lacks semantic and global shape constraints, resulting in artifacts~\cite{levi2014smooth} under part-level edits. We address this with a two-stage pipeline (Fig.~\ref{fig:placeholder}): \textit{(i) locally rigid 3D editing} via CS-ARAP, followed by \textit{(ii) global refinement} guided by a pre-trained diffusion model~\cite{song2022denoisingdiffusionimplicitmodels}.

\noindent\textbf{Mapping user edits.} Given $H_{pose} = [R_{ip} \mid T_{ip}]$, where $R_{ip} \in SO(3)$ and $T_{ip} \in \mathbb{R}^3$, and keypoint handles $V_{handles}$, we transform user edits onto $\mathcal{M}_{obj} = (V,F)$. Each handle $p_{h} \in V_{handles}$ is transformed about the geometric centroid $p_{c}$:
\begin{align}
    V_{tar} = R_{ip}(p_{h} - p_{c}) + T_{ip},
\end{align}
yielding target vertices $V_{tar}$. Handle vertices $v_{h} \in V_{handles}$ are movable and minimize the ARAP energy to preserve local rigidity; all remaining vertices are fixed.

\noindent\textbf{Locally rigid 3D editing.}
Classical ARAP~\cite{sorkine2007rigid} solves for rotations based on local rigidity alone, producing $C^0$ rotation fields that can contain high-frequency spikes at constraint boundaries, particularly with fine mesh resolution and few-point handles (Fig.~\ref{fig:registration_ablation}). The standard remedy is SR-ARAP~\cite{levi2014smooth}, which augments the energy with $\alpha \sum_{i} \sum_{j \in \mathcal{N}_i} \|R_i - R_j\|^2$. However, this requires jointly optimizing an additional term at every iteration, increasing per-iteration cost, and does not resolve positional kinks at constraint boundaries.

\begin{figure*}[!t]
  \centering
  \setlength{\tabcolsep}{0.25pt}
  \includegraphics[width=\textwidth]{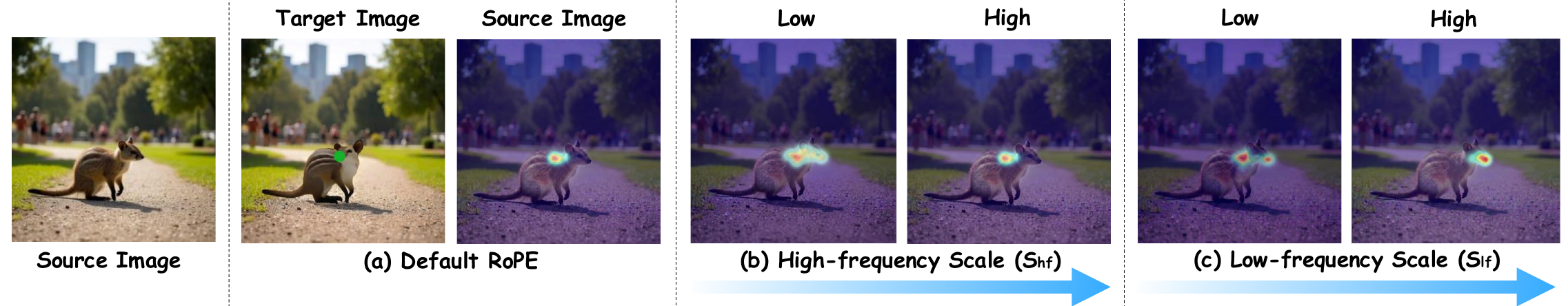}
  \caption{\textbf{Frequency-aware modulation of cross-image attention.} For a
query in the edited target (green marker), we visualize its attention over the
source. Default RoPE attends to the part's pre-edit location (a); this bias vanishes when high-frequency  ($S_{hf}$) chunks are removed (b) and attention shifts to semantically matching regions as low-frequency chunks ($S_{lf}$) are amplified (c)  Details in supp.}
  \label{fig:heat_map}
\end{figure*}

We propose \textbf{CS-ARAP}, which achieves smooth rotation fields without auxiliary optimization. Rather than penalizing rotation differences post-hoc, we smooth the \textit{input} to the rotation projection: we replace the per-vertex covariance $S_i$ with a neighborhood-averaged $\tilde{S}_i$ before SVD. This acts as a low-pass filter directly on the affine transformations that determine $R_i$, producing spatially coherent rotations. We compute $\tilde{S}_i$ via a convex combination of the local covariance and the neighborhood average:
\begin{equation}
    \tilde{S}_{i} = (1-\beta)S_{i} + \beta \left( \frac{1}{|\mathcal{N}_{i}|}\sum_{j\in \mathcal{N}_{i}}S_{j} \right),
\end{equation}
where $\beta \in [0, 1]$ controls smoothing strength. We obtain $\tilde{R}_i$ by projecting $\tilde{S}_i$ via SVD, similar to~\cite{sorkine2007rigid}. By modifying the input to the standard SVD-based rotation projection rather than introducing a separate rotation-smoothness optimization~\cite{levi2014smooth}, we avoid the additional coupled rotation optimization of SR-ARAP while encouraging spatial coherence. In practice, CS-ARAP also converges substantially faster than ARAP and SR-ARAP, as quantified in Table~\ref{tab:solver_efficiency}. However, rotation smoothing alone is insufficient: the standard ARAP position solve uses only the second-order Laplacian, which can produce positional deformities at constraint boundaries. We thus augment it with a bi-Laplacian term, which penalizes large curvature changes:
\begin{equation}
\begin{aligned}
L_{\text{coarse}}
&= (1-\lambda)E_{\text{ARAP}} + \lambda E_{\text{blp}}, \\
E_{\text{blp}}
&= \sum_i \left\| (LV')_i - \tilde{R}_i (LV)_i \right\|^2.
\end{aligned}
\end{equation}
Here $(LV)_i$ denotes the discrete Laplacian at vertex $i$ (Sec.~\ref{sec:preliminary}). Minimizing $L_{\text{coarse}}$ with $\tilde{R}$ fixed yields:
\begin{equation}
    \left((1-\lambda)L + \lambda L M^{-1} L\right) V' = b,
\end{equation}
where $b$ is constructed from the smoothed rotations and $M$ is the mass matrix. We obtain coarse mesh $M_{obj}^{'} = (V^{'},F)$. This higher-order smoothness is particularly beneficial for part-level editing (Fig.~\ref{fig:registration_ablation}): handles manipulate localized regions while preserving smooth deformation across boundaries. While $L_{\text{coarse}}$ ensures geometric smoothness, it has no semantic understanding of the object's shape, so large edits can yield smooth but unrealistic global deformation.

\noindent\textbf{Global Refinement.} To optimize for global mesh deformations, we initialize the Jacobian field $\mathcal{J}$ from Stage 1 vertices $V'$ as $J_f = \nabla_f V'$ and optimize it to balance geometric constraints with semantic plausibility. To preserve visual identity, we condition a pre-trained Stable Diffusion model using IP-Adapter~\cite{ye2023ip}, injecting CLIP image features from renderings of the original mesh $\mathcal{M}_{obj}$ via cross-attention. The conditioned model $\epsilon_\phi$ serves as the critic. At each iteration, we render the deformed mesh $\hat{\mathcal{M}}_{obj} = (V^*, F)$ from a randomly sampled viewpoint using a differentiable renderer $\mathcal{R}$, producing $I_{\text{render}} = \mathcal{R}(\hat{\mathcal{M}}_{obj}, C)$. We update $\mathcal{J}$ via gradient descent to minimize:
\begin{equation}
\label{eq:sds}
\begin{aligned}
L_{\text{refine}}
&= \mathcal{L}_{\text{handle}}
+ \omega_{\text{SDS}} \mathcal{L}_{\text{SDS}}, \\
&\text{where}\quad
\mathcal{L}_{\text{handle}}
= \left\| K_h \mathcal{V}(\mathcal{J}) - T_h \right\|^2.
\end{aligned}
\end{equation}
Here $K_h \in \{0,1\}^{V_h \times V}$ is the handle selection matrix and $\mathcal{V}(\mathcal{J})$ denotes the vertex positions recovered from $\mathcal{J}$ via the differentiable Poisson solve (Sec.~\ref{sec:preliminary}). The handle constraint enforces the user's editing instructions by minimizing the distance between handle vertices and their targets. The $\mathcal{L}_{\text{SDS}}$ term provides gradients that push the rendered geometry toward the distribution of plausible images defined by $\epsilon_\phi$, penalizing unrealistic distortions while preserving the object's learned identity.

\vspace{-4pt}

\subsection{Photorealistic Appearance Blending}
\label{sec:tex_trans_sec}


We composite the edited render into the background to restore fine-grained appearance while strictly preserving geometry. To avoid the geometric perturbation and semantic leakage inherent to inversion-based editing~\cite{cao2023masactrl,yenphraphai2024image,alaluf2023crossimage}, we inject appearance through an inversion-free attention formulation. Because standard U-Net attention sharing~\cite{hertz2024style} fails in position-encoded DiTs, we introduce a novel frequency-aware attention modulation to accurately transfer the source image's appearance.

\noindent \textbf{Scene Blending and Transformation.}
We render $\hat{\mathcal{M}_{3D}}$ as $I_{render}$ using camera intrinsics $H_{obj}$ and correct minor scale and positional discrepancies between $I$ and $I_{render}$ via a transformation $\pi(\cdot)$ on camera translation and scale. However, two issues remain: \textit{(i)} compositing artifacts from geometric edits, and \textit{(ii)} inaccurate textures from missing material and illumination modeling in 3D lifting. We address (i) by removing the original object via off-the-shelf inpainting~\cite{brushnet} to obtain a consistent background $I_{bg}$, then compositing using alpha blending:
\begin{equation}
\label{eq:blend}
I_{g} = \mathbb{B}(\pi(I_{render}), I_{bg}),
\end{equation}
where $\mathbb{B}(\cdot)$ denotes the blending operator. The result preserves geometry but lacks photorealism due to inaccurate appearance and texture. Instead of explicit relighting, we transfer the object appearance from source image $I$.
\begin{table*}[!t]
\footnotesize
\centering
\caption{\textbf{Quantitative comparison against state-of-the-art methods on GeoBench.} RADI3N achieves the best geometric fidelity and appearance consistency across nearly all metrics, while maintaining competitive appearance and identity preservation.
}
\label{tab:results}
\resizebox{\textwidth}{!}{
\begin{tabular}{l cccc cc ccc}
\toprule
{\textbf{Method}} & \multicolumn{4}{c}{\textbf{Geometric Fidelity}} & \multicolumn{2}{c}{\textbf{Appearance \& Identity}} & \multicolumn{3}{c}{\textbf{Consistency}} \\
\cmidrule(lr){2-5} \cmidrule(lr){6-7} \cmidrule(lr){8-10}
& WE $\downarrow$ & D-RMSE $\downarrow$ & MAAE$^{\circ}$ $\downarrow$ & Acc@5$^{\circ}$ $\uparrow$ & CLIP-I $\uparrow$ & FID $\downarrow$ & SUBC $\uparrow$ & DINOv2 $\downarrow$ & BC $\uparrow$ \\
\midrule
GoodDrag~\cite{zhang2024gooddrag}        & 0.086 & 2.77 & 26.84 & 0.084 & 0.894 & 167.90 & 0.797 & 1913.67 & 0.897 \\
Image Sculpting~\cite{yenphraphai2024image} & 0.082 & 2.75 & 19.83 & 0.293 & 0.897 & 160.46 & 0.808 &1909.27 & 0.894 \\
FlowDrag~\cite{koo2025flowdrag}        & 0.077 & \underline{2.14} & 21.45 & 0.176 & 0.901 & 150.23 & 0.809 & 1901.10 & 0.933 \\
FreeFine~\cite{zhu2025training}        & 0.114 & 2.51 & 20.31 & 0.285 & 0.897 & 152.37 & 0.814 & 1901.23 & 0.821 \\
Qwen-Img-Edit~\cite{qwen_image_edit}   & \underline{0.076} & 2.23 & 18.47 & 0.241 & \underline{0.929} & \underline{147.76} & 0.826 & 1896.16 & \textbf{0.952} \\

Nano Banana 2~\cite{nanobanana}   & 0.081 & 2.17 & \underline{12.04} & \underline{0.374} & \textbf{0.933} & \textbf{144.08} & \underline{0.827} & \underline{1890.05} & 0.938 \\
\midrule
\rowcolor[gray]{0.9}
\textbf{RADI3N}   & \textbf{0.067} & \textbf{1.91} & \textbf{7.07} & \textbf{0.789} & 0.924 & 148.47 & \textbf{0.835} & \textbf{1874.42} & \underline{0.949} \\
\bottomrule
\end{tabular}
}
\end{table*}

\noindent\textbf{Appearance Injection.}
Given the source image $I$ and the initial composite $I_g$, we extract spatial latents patchified into 1D token sequences, yielding target and source tokens ($\mathcal{T}^{\mathrm{tar}}$, $\mathcal{T}^{\mathrm{so}}$). Joint transformer blocks linearly project these into queries ($Q$), keys ($K$), and values ($V$). To inject source appearance
into the edited geometry without relying on costly inversion, which is prone to geometric drift and semantic leakage~\cite{alaluf2023crossimage,cao2023masactrl}, we employ an inversion-free attention-sharing formulation~\cite{hertz2024style}. Target key and value sequences are extended by concatenating the corresponding source keys and values. Letting $\oplus$ denote concatenation along the sequence dimension:
\begin{align}
Q &= \mathrm{RoPE}\big(\big[\tilde{Q}^{\mathrm{img}}_{\mathrm{tar}} \oplus Q^{\mathrm{txt}}\big]\big), \nonumber\\
K &= \mathrm{RoPE}\big(\big[{K}^{\mathrm{img}}_{\mathrm{tar}} \oplus K^{\mathrm{txt}}\big]\big) \oplus s \cdot \mathrm{RoPE}\big(K^{\mathrm{img}}_{\mathrm{so}}\big), \label{eq:ext_kv}\\
V &= \big[V^{\mathrm{img}}_{\mathrm{tar}} \oplus V^{\mathrm{txt}}\big] \oplus V^{\mathrm{img}}_{\mathrm{so}}, \nonumber
\end{align}
where $\tilde{Q}_{\mathrm{tar}} = \text{AdaIn}(Q_{\mathrm{tar}},Q_{\mathrm{so}})$ and $Q^{\mathrm{txt}}/K^{\mathrm{txt}}/V^{\mathrm{txt}}$ denotes the textual prompt tokens. The target queries then attend to this extended context to compute the attention output $O_{\mathrm{tar}}$. While this lets queries over the edited geometry ($Q_{\mathrm{tar}}$) draw appearance features directly from the source values ($V_{\mathrm{so}}$), the Rotary Positional Embeddings (RoPE) in DiTs introduce a uniquely harmful bias. RoPE's high-frequency components cause shared-attention logits to be dominated by relative token positions, forcing target queries to favor spatially aligned source tokens over semantic matches~\cite{su2024roformer,zhang2026alignedgen}. Since $I$ and $I_g$ depict the same scene, the spatially aligned source token for an edited part is its \emph{pre-edit} texture , the exact location appearance must not be copied from. Consequently, this positional attention repaints the part at its stale coordinates (Fig.~\ref{fig:heat_map}a).

\noindent\textbf{Modulated Texture Transfer.}
To resolve this positional bias and transfer the correct appearance, we leverage the non-uniform nature of RoPE's spectrum. RoPE~\cite{su2024roformer} rotates embeddings in 2D chunks at geometrically decaying frequencies: high-frequency chunks decay steeply with displacement to cause the aforementioned spatial pinning, while low-frequency chunks remain largely displacement-invariant to support semantic matching. For editing, the reference-copying bias lives entirely in the high frequencies (Fig.~\ref{fig:heat_map}b). Conversely, low frequencies enable semantic matching, letting appearance seamlessly follow the part to its new pose (Fig.~\ref{fig:heat_map}c). Because the edited geometry is already determined by the rendered composite, any residual high-frequency source contribution would incorrectly re-introduce attention to pre-edit positions. We therefore discard the high-frequency components of RoPE by replacing the scalar gain $s$ in Eq.~\ref{eq:ext_kv} with a low-pass mask over the source keys:
\begin{equation}
\hat{K}^{\mathrm{img}}_{\mathrm{so}} = M_{\mathrm{low}}(S_{lf}) \odot \mathrm{RoPE}\big(K^{\mathrm{img}}_{\mathrm{so}}\big). \label{eq:lp_key}
\end{equation}
This mask $M_{\mathrm{low}}(S_{lf})$ suppresses high-frequency chunks to eliminate strict positional bias and structural leakage. Simultaneously, it scales the retained low-frequency chunks by a strength $s_{lf} \ge 1$ (applied independently to the $x$- and $y$-axis partitions) to strengthen coarse regional priors. By modulating only the source image keys, attention is resolved via semantic correspondence within the correct coarse region. This allows low-spatial-frequency cues like texture and illumination to accurately follow the part onto its new pose without warping the edited geometry.

\begin{figure*}[!htbp]
  \centering
  \setlength{\tabcolsep}{0.25pt}
  \includegraphics[width=\textwidth]{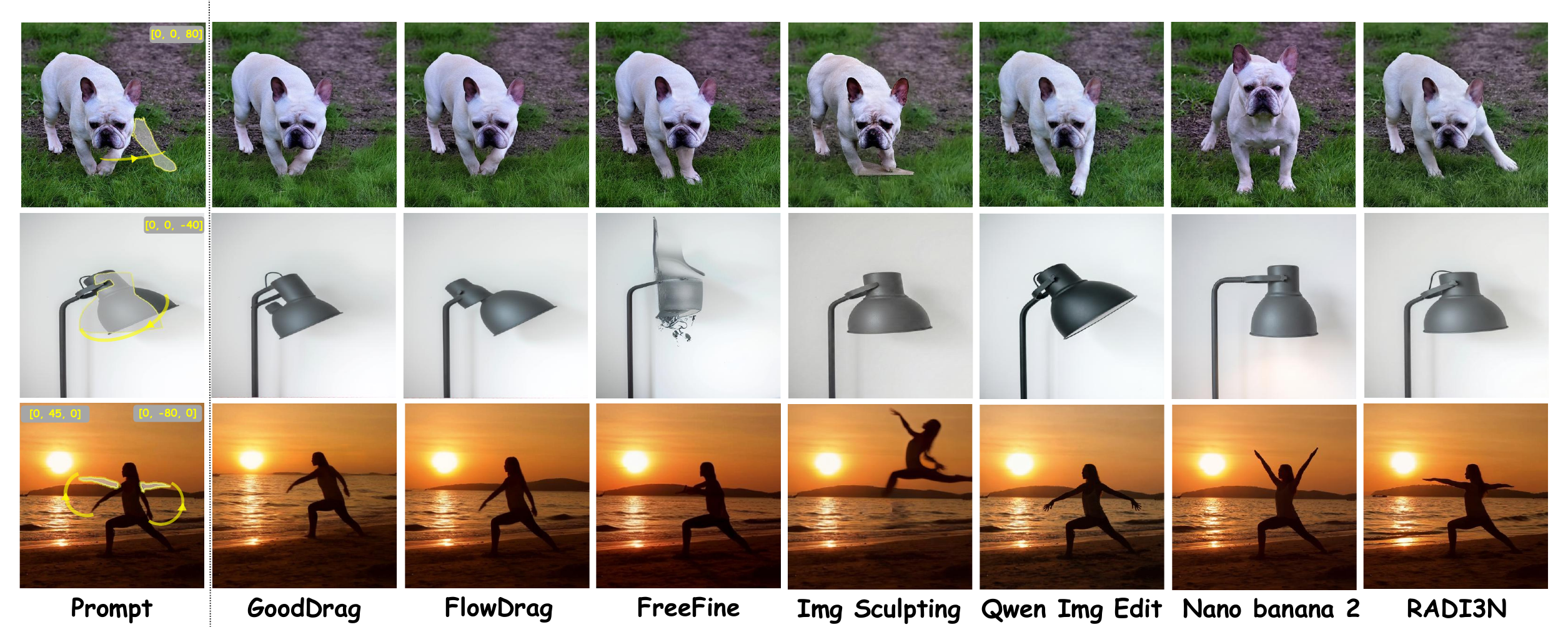} \\
  \caption{\textbf{Qualitative comparison with state-of-the-art methods.}
We compare RADI3N against other state-of-the-art methods across diverse object categories. Our method consistently produces
geometrically faithful and photorealistic edits, while baselines
exhibit artifacts, implausible geometry, or loss of scene context.}
  \label{fig:method_comparison}
\end{figure*}

\vspace{-4pt}
\section{Experiments}
\label{experiments}

\subsection{Datasets and Evaluation Metrics}

RADI3N is evaluated on DragBench (205 images, 349 handle-target pairs) and GeoBench (811 source images, 5,988 object-centric editing instructions spanning move, rotate, and resize). We assess performance along three axes.

\begin{figure*}[!t]
  \centering
  \setlength{\tabcolsep}{0.25pt}
  \includegraphics[width=\textwidth]{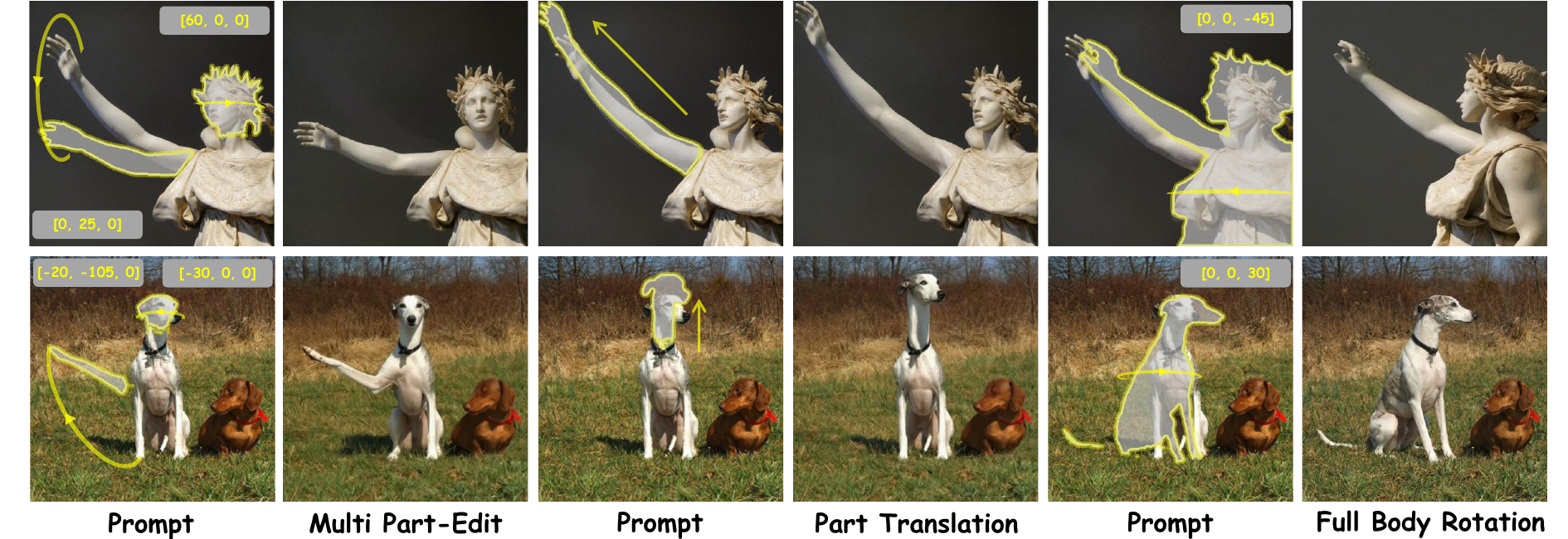}
  \caption{\textbf{Qualitative results across editing operations.}
RADI3N supports diverse 6DOF control modes, including
multi-part editing, part translation, and full-body
rotation (F.B.R).}
  \label{fig:qualitative_samples}
\end{figure*}

\begin{figure}[t]
    \centering
    \begin{subfigure}[b]{\linewidth}
        \centering
        \includegraphics[width=\linewidth]{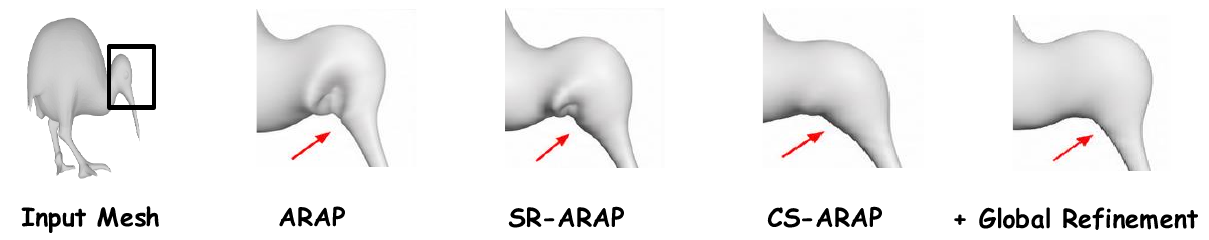}
        \caption{Comparison of deformation methods}
    \end{subfigure}

    \caption{We compare ARAP and SR-ARAP with ours.}
    \label{fig:registration_ablation}
\end{figure}
\begin{table}[t]
    \centering
    \caption{\textbf{Component Ablations and Efficiency Analysis.}}
    \label{tab:combined_quantitative}
    \scriptsize
    \setlength{\tabcolsep}{3.5pt}
    \renewcommand{\arraystretch}{1.05}


    \begin{subtable}[t]{0.48\columnwidth}
        \centering
        \caption{Geometric Fidelity}
        \label{tab:ablation_geometry}
        \resizebox{\linewidth}{!}{
        \begin{tabular}{lcc}
            \toprule
            \textbf{Configuration}
            & \textbf{WE $\downarrow$}
            & \textbf{D-RMSE $\downarrow$} \\
            \midrule
            ARAP~\cite{sorkine2007rigid}
            & 0.147 & 2.39 \\
            SR-ARAP~\cite{levi2014smooth}
            & 0.122 & 2.30 \\
            ARAP+SDS
            & 0.105 & 2.40 \\
            CS-ARAP
            & \underline{0.091} & \underline{2.16} \\
            \midrule
            \rowcolor[gray]{0.9}
            \textbf{RADI3N}
            & \textbf{0.067} & \textbf{1.91} \\
            \bottomrule
        \end{tabular}
        }
    \end{subtable}
    \hfill
    \begin{subtable}[t]{0.48\columnwidth}
        \centering
        \caption{Attention Modulation}
        \label{tab:frequency_ablation}
        \resizebox{\linewidth}{!}{
       \begin{tabular}{lcc}
    \toprule
    \textbf{Configuration}
    & \textbf{CLIP-I $\uparrow$}
    & \textbf{WE $\downarrow$} \\
    \midrule
    Default RoPE
    & 0.895 & 0.079 \\
    + AlignedGen~\cite{zhang2026alignedgen}
    & 0.904 & 0.073 \\
    Only high-frequencies
    & 0.891 & 0.076 \\
    high and low
    & 0.911 & 0.070 \\
    \midrule
    \rowcolor[gray]{0.9}
    \textbf{Only low-frequencies}
    & \textbf{0.924} & \textbf{0.067} \\
    \bottomrule
\end{tabular}
}
    \end{subtable}




    \begin{subtable}[t]{0.48\columnwidth}
        \centering
        \caption{Solver Convergence}
        \label{tab:solver_efficiency}
        \resizebox{\linewidth}{!}{
        \begin{tabular}{lcc}
            \toprule
            \textbf{Method}
            & \textbf{Solve (s) $\downarrow$}
            & \textbf{Iter. $\downarrow$} \\
            \midrule
            ARAP~\cite{sorkine2007rigid}
            & \underline{2.18} & \underline{104} \\
            SR-ARAP~\cite{levi2014smooth}
            & 2.81 & 187 \\
            \midrule
            \rowcolor[gray]{0.9}
            \textbf{CS-ARAP}
            & \textbf{1.79} & \textbf{69} \\
            \bottomrule
        \end{tabular}
        }
    \end{subtable}
   \hfill
    \begin{subtable}[t]{0.48\columnwidth}
        \centering
        \caption{End-to-End Efficiency}
        \label{tab:timing_comparison}
        \resizebox{\linewidth}{!}{
        \begin{tabular}{lcc}
            \toprule
            \textbf{Method}
            & \textbf{Time (s) $\downarrow$}
            & \textbf{D-RMSE $\downarrow$} \\
            \midrule
            Img Sculpting~\cite{yenphraphai2024image}
            & 450--500 & 2.75 \\
            FlowDrag~\cite{koo2025flowdrag}
            & 120--135 & \underline{2.14} \\
            FreeFine~\cite{zhu2025training}
            & \textbf{40--45} & 2.51 \\
            \midrule
            \rowcolor[gray]{0.9}
            \textbf{RADI3N}
            & \underline{95--100} & \textbf{1.91} \\
            \bottomrule
        \end{tabular}
        }
    \end{subtable}
\end{table}




\begin{figure*}[!t]
    \centering

    \begin{subfigure}[t]{0.28\textwidth}
        \centering
        \includegraphics[width=\linewidth]{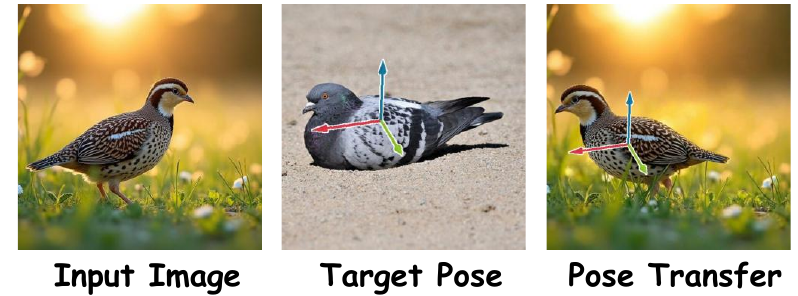}
        \caption{Pose Transfer}
        \label{fig:pose_transfer}
    \end{subfigure}%
    \hfill
    \begin{subfigure}[t]{0.41\textwidth}
        \centering
        \includegraphics[width=\linewidth]{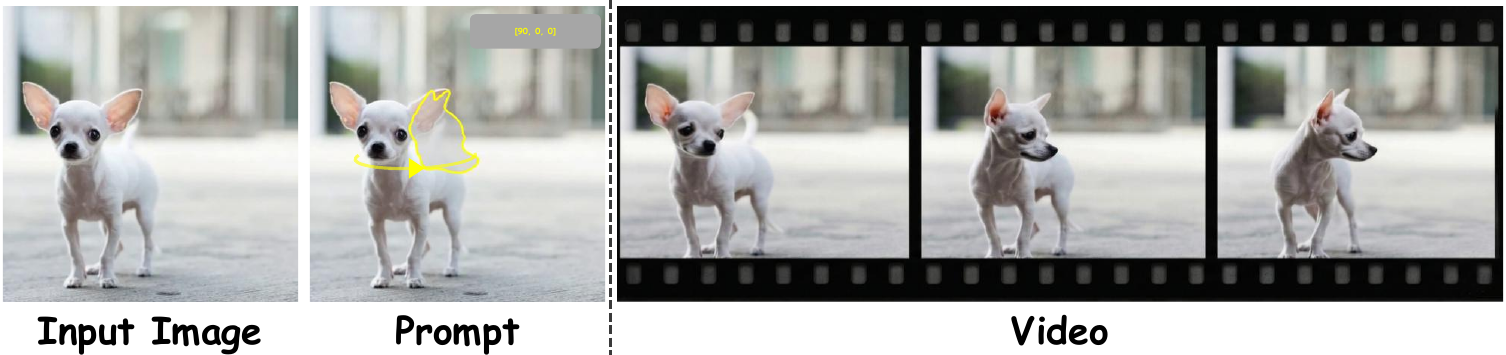}
        \caption{Video Generation}
        \label{fig:gif_down_stream}
    \end{subfigure}%
    \hfill
    \begin{subfigure}[t]{0.28\textwidth}
        \centering
        \includegraphics[width=\linewidth]{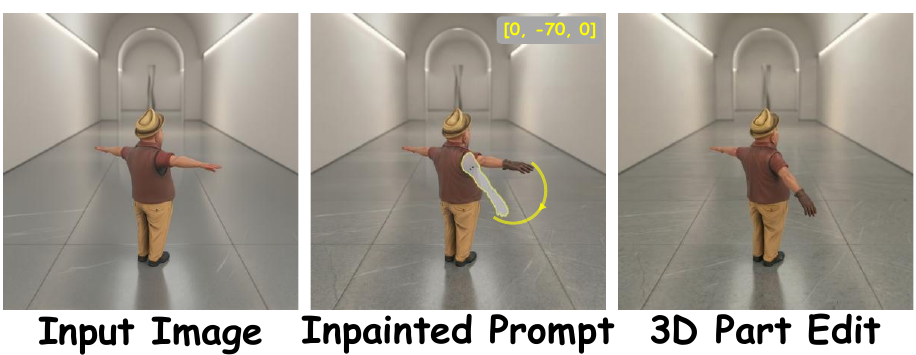}
        \caption{3D Inpainting}
        \label{fig:downstream_img_inpaint}
    \end{subfigure}
\caption{\textbf{Applications of RADI3N.} We demonstrate the usability across several tasks. Details in supp.}
\label{fig:applications}
\end{figure*}

For \textbf{geometric fidelity}, we report \textbf{Warp Error (WE)}~\cite{sajnani2025geodiffuser} and \textbf{D-RMSE}~\cite{yenphraphai2024image}. For \textbf{angular accuracy}, we lift edited images to 3D via SAM3D, recover the applied rotation, and report \textbf{MAAE} across azimuth, polar, and rotation axes along with \textbf{Acc@5$^{\circ}$}. To rule out evaluation circularity from using SAM3D as both editing backbone and angular evaluator, we re-run the angular evaluation using Hunyuan3D~\cite{zhao2025hunyuan3d} as an independent lifter (ref. supp.). For \textbf{identity preservation} we use \textbf{CLIP-I}~\cite{radford2021learning} and \textbf{FID}~\cite{heusel2017gans}; for \textbf{contextual consistency} we report \textbf{Subject Consistency (SUBC)} and \textbf{Background Consistency (BC)}~\cite{huang2024vbench}, complemented by \textbf{DINOv2}~\cite{dinov2} feature distances.




\vspace{-5pt}
\subsection{Results}

\noindent\textbf{Qualitative Results}
Fig.~\ref{fig:method_comparison} compares representative methods. Drag-based approaches~\cite{zhang2024gooddrag,koo2025flowdrag} and FreeFine often struggle with part articulation and foreground morphing under large deformations. Qwen~\cite{qwen_image_edit} and Nano Banana 2 \cite{nanobanana} preserve appearance but frequently fail to execute the intended transformation. Image Sculpting handles moderate edits but introduces blending inconsistencies and degrades during texture transfer. RADI3N consistently achieves geometrically accurate edits while preserving subject identity and background coherence. Diverse edits in Fig.~\ref{fig:qualitative_samples}.

\noindent\textbf{Quantitative Results.} Table~\ref{tab:results} shows that RADI3N achieves the best geometric fidelity (WE 0.067, D-RMSE 1.91) and angular accuracy (MAAE 7.07$^\circ$, Acc@5$^\circ$ 78.9\%). It reduces MAAE by 41\% over the strongest baseline, Nano Banana 2 (12.04$^\circ$), while more than doubling Acc@5$^\circ$ (37.4\%). Nano Banana 2 achieves the best CLIP-I (0.933) and FID (144.08), followed by Qwen-Img-Edit (0.929, 147.76), while RADI3N remains competitive (0.924, 148.47) and achieves the best SUBC (0.835) and DINOv2 distance (1874.42). Qwen-Img-Edit has the highest BC (0.952), with RADI3N close behind (0.949), indicating strong background preservation. Extended comparisons and DragBench results are provided in the supplementary.

\subsection{Ablation Studies}
\noindent\textbf{Geometric Fidelity}
Table~\ref{tab:ablation_geometry} evaluates each component. CS-ARAP improves
over ARAP via covariance smoothing and bi-Laplacian solving,
removing the surface artifacts in Fig.~\ref{fig:registration_ablation}. SDS on
standard ARAP performs \emph{worse} than CS-ARAP alone, showing that coarse
initialization quality bounds what SDS can recover; the full pipeline is best
(WE 0.067, D-RMSE 1.91), as refinement on a CS-ARAP initialization corrects
residual inconsistencies.

\noindent\textbf{Model Efficiency}
RADI3N attains the best D-RMSE at 95-100s per edit, $1.3\times$ faster than
FlowDrag and ${\sim}5\times$ faster than Image Sculpting
(Table~\ref{tab:timing_comparison}). FreeFine is faster but lags on geometric
fidelity (D-RMSE 2.51 vs.\ 1.91), reflecting the trade-off between feed-forward
speed and optimization for 3D-consistent edits. Covariance smoothing also
accelerates convergence: CS-ARAP needs 69 iterations against 104 (ARAP) and 187
(SR-ARAP), the lowest overall per-iteration cost and solve time (Table~\ref{tab:solver_efficiency}).

\noindent\textbf{Attention Modulation}
Table~\ref{tab:frequency_ablation} ablates our modulation against Default RoPE and AlignedGen~\cite{zhang2026alignedgen}, which reindexes source coordinates to counter positional bias. Retaining only low-frequency components performs best, confirming that high-frequency chunks carry the positional bias while low-frequency chunks drive correct semantic transfer. Fig.~\ref{fig:heat_map} shows the corresponding attention shift.


\noindent\textbf{User Study}
34 participants judged 35 edited images, selecting the result best reflecting
the intended transformation while remaining realistic within the masked region.
RADI3N is preferred in \textbf{79.6\%} of comparisons, above all baselines.

\subsection{Applications}
\vspace{-5pt}
Beyond local edits, our framework enables several downstream applications (Fig.~\ref{fig:applications}). For \textbf{pose transfer}, we estimate poses with Orient Anything~\cite{ori} and apply the relative rigid transform to the mesh, preserving source appearance in the target pose. For \textbf{video generation}, Blender-interpolated mesh deformations combined with Diffusion as Shader~\cite{gu2025diffusion} yield photorealistic tracking videos from a single rendered frame. For \textbf{3D-aware inpainting}, EditP23~\cite{bar2025editp23}-inserted objects pass directly through our pipeline, preserving content and geometric consistency under large transformations.



\vspace{-5pt}
\section{Conclusion}
\label{conclusion}

We present \textbf{RADI3N}, a framework for semantically-grounded 3D part-level image editing from a single image. RADI3N couples semantic part understanding with robust geometric deformation and frequency-modulated texture transfer in a unified pipeline, enabling precise 6-DOF edits without per-object rigging or task-specific training, and achieving state-of-the-art geometric fidelity and identity preservation on DragBench and GeoBench.
Our design has some limitations: CS-ARAP's local rigidity constraint can restrict extreme non-rigid articulation, and applying a uniform frequency threshold can occasionally leave minor blending seams under severe viewpoint shifts. We detail these cases in the supplementary, with adaptive frequency control and non-rigid modeling as future directions.



{
    \small
    \bibliographystyle{ieeenat_fullname}
    \bibliography{main}
}

\clearpage
\makeatletter
\setlength{\@dblfptop}{0pt}
\setlength{\@dblfpsep}{12pt}
\setlength{\@dblfpbot}{0pt plus 1fil}
\setlength{\@fptop}{0pt}
\setlength{\@fpsep}{12pt}
\setlength{\@fpbot}{0pt plus 1fil}
\makeatother
\maketitlesupplementary
\appendix
\section{Implementation Details}
\subsection{Geometric Editing}
\label{sec:impl_geo}

Our geometric part-editing pipeline runs in two stages. In Stage~1, the CS-ARAP
deformation is initialized with a bi-Laplacian weight
$\lambda=0.95$ and covariance smoothing $\beta \in [0.0, 0.2]$, and the
alternating minimization (Alg.~\ref{alg:smooth-arap}) is iterated until the relative
change in vertex positions falls below $\epsilon = 1\times10^{-4}$. Input
meshes contain $\approx\!30$k vertices on average and are driven by
$|\mathcal{B}| \in [1, 20]$ point handles.
We set $\lambda$ high to heavily favor $C^2$ positional smoothness,
producing a spike-free mesh that serves as a stable initialization
for Stage~2. The smoothing parameter $\beta$ is set proportional
to edit magnitude: small handle displacements use $\beta=0$, while
large articulations require $\beta=0.2$ to prevent rotation
discontinuities across the surface.

In Stage~2, we use Stable Diffusion v2.1-base as the diffusion
prior, conditioned via IP-Adapter, which injects
CLIP image features from renderings of the source mesh to preserve
object identity without per-instance fine-tuning. The Jacobian field is optimized using Adam at a learning
rate of $1 \times 10^{-3}$ for $N_{iter} = 1{,}000$ iterations. Refinement draws
from a fixed candidate camera set $\mathcal{C}$ of $|\mathcal{C}| = 20$ poses:
10 azimuths spanning $360^\circ$, each at two elevations, $-10^\circ$ and
$+15^\circ$. The azimuth grid is rotated once at initialization so that one
azimuth aligns with the estimated object orientation, and is held fixed
thereafter. At each iteration we draw a single camera
$C \sim \mathrm{Uniform}(\mathcal{C})$ and render at $512 \times 512$, so each
pose is visited roughly 50 times over the course of refinement. SDS uses a CFG
scale of 100.0, is weighted by $\omega_{\mathrm{SDS}} = 10$ (Eq.~9), and samples
timesteps uniformly from $t \in [0.02, 0.98]$.

\subsection{Appearance Blending}
\label{sec:impl_app}

Background completion uses BrushNet to
inpaint the masked region left by the removed object. 

\paragraph{Backbone and sampling.}
Appearance transfer runs on FLUX.1-dev at $1024\times1024$ with the
rectified-flow sampler, 30 sampling steps, and guidance scale 3.5. Neither
image is inverted: the source $I$ and the composite $I_g$ are both
forward-noised to a shared timestep (strength 0.6), and the substitution of
Eq.~(11) is active from step 4 of 30 onward, applied to the single-stream blocks
(layers $[19,57)$ under consecutive indexing of the 19 double-stream and 38
single-stream blocks; the double-stream blocks are left unmodified) that prior
work identifies as governing appearance and fine detail. Only
source image keys and values are concatenated, so source text tokens do not
affect the target prompt conditioning. Both branches are conditioned on the same
text prompt, a short caption of the source object; the edit instruction is not
passed to FLUX, since the target geometry is already fixed by the composite
$I_g$. Source token positions are left unshifted, so the modulation of
Eq.~(12) is the only positional intervention.

\noindent\textbf{Frequency-modulated source keys.} The mask
$\mathcal{M}_{low}(S_{lf})$ of Eq.~(12) replaces the scalar gain $s$ of
Eq.~(11) with a per-channel-pair rescaling of the source image keys.
FLUX partitions its $D = 128$ channels as $(16, 56, 56)$ over the
temporal, $y$- and $x$-axes, giving $n = 28$ channel pairs per spatial axis.
Writing $\tilde{d} = d/n$ for the normalized pair index, where early pairs
carry the highest frequency, the per-pair scale is
\begin{equation}
    s_d = S_{lf}\, \tilde{d}^{\,k},
    \label{eq:ramp}
\end{equation}
with the two channels of each cos/sin pair sharing a scale. The scale
vanishes at the highest-frequency pair and rises monotonically toward
$S_{lf}$ at the lowest, giving a smooth low-pass response: the
high-frequency band is attenuated to a negligible fraction of its original
magnitude while the low-frequency band is amplified. We set $k = 2.0$ and a
single $S_{lf} = 2$ shared by both spatial axes, held constant across
denoising steps; larger values yield no further improvement. The ramp is
applied independently to the $x$- and $y$-axis channel partitions, and the
unrotated temporal partition is scaled by $S_{lf}$ throughout. Since pairs are indexed from $d = 0$ at the highest frequency, the
suppression the main paper describes as discarding is exact at the top of the
spectrum ($s_0 = 0$) and near-exact below it, and is realized as a monotone
ramp rather than a binary cutoff; the \emph{uniform frequency threshold} of the
main paper's limitations refers to holding $k$ and $S_{lf}$ fixed across
images, axes and denoising steps. 
\section{RoPE and Shared Attention}
\label{sec:rope_prelim}

We recall the construction of rotary positional embeddings and use it to
show why naive attention sharing fails on a RoPE-based backbone, motivating
the modulation of Eq.~(12).

\subsection{Rotary Positional Encoding}

Let $q \in \mathbb{R}^{D}$ be a query at position $m$ and
$k \in \mathbb{R}^{D}$ a key at position $n$. RoPE
partitions each vector into $D/2$ two-dimensional \emph{chunks}, indexed by
$d$, and rotates each independently:

\begin{figure*}[!htbp]
\centering
\includegraphics[width=\linewidth]{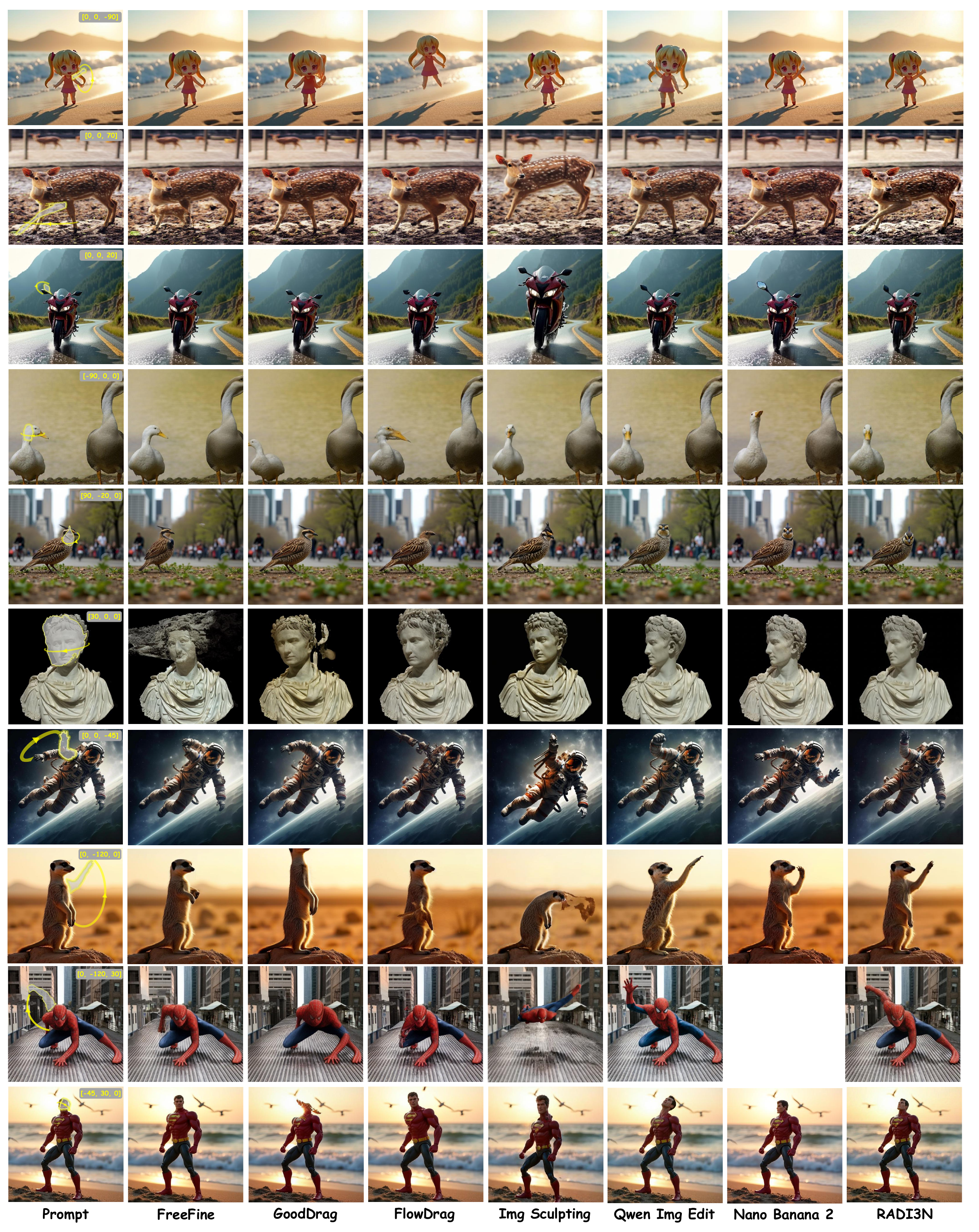}
\caption{Extended qualitative comparisons across additional
object categories. Each row shows the input, editing prompt,
and outputs from all methods.}
\label{fig:fig5_extended}
\end{figure*}
\begin{equation}
    \hat{q}_d = R_{m\theta_d}\, q_d,
    \qquad
    \hat{k}_d = R_{n\theta_d}\, k_d,
\end{equation}
where $R_\phi$ is the $2\!\times\!2$ rotation by $\phi$ and the per-chunk
frequency follows a geometric series,
\begin{equation}
    \theta_d = \theta_{\text{base}}^{\,2d/D},
    \qquad
    d \in \left\{0, \dots, \tfrac{D}{2}-1\right\},
    \label{eq:rope_freq}
\end{equation}
so that early chunks carry the highest frequencies. Because rotations
compose, the attention logit depends only on the relative displacement
$\Delta = n - m$:
\begin{equation}
    \langle \hat{q}, \hat{k} \rangle
    = \sum_{d} q_d^{\top} R_{m\theta_d}^{\top} R_{n\theta_d} k_d
    = \sum_{d} \big\langle q_d,\, R_{\Delta\theta_d}\, k_d \big\rangle .
    \label{eq:rope_rel}
\end{equation}
For images, each token is split into sub-vectors encoding the $x$- and
$y$-axes, so that $\Delta$ is evaluated per axis on the corresponding
chunks.

\subsection{Why Naive Attention Sharing Fails}
\label{sec:naive_sharing_fails}

Writing each chunk of Eq.~\eqref{eq:rope_rel} in polar form, the rotation
becomes an additive phase shift:
\begin{equation}
    \langle \hat{q}_d, \hat{k}_d \rangle
    = \| q_d \| \, \| k_d \| \cos\!\left( \alpha_d + \Delta\theta_d \right),
    \label{eq:rope_polar}
\end{equation}
where $\alpha_d$ is the angle between the unrotated chunks. Content
similarity and position therefore enter additively inside a single cosine.

Extending shared attention to a RoPE-based backbone
concatenates source keys and values onto the target sequence (Eq.~11). The
two token sets occupy the same $h \times w$ latent grid, so a source and a
target token at identical coordinates have $\Delta = 0$ between
them. Substituting into
Eq.~\eqref{eq:rope_polar}, the positional phase vanishes in every chunk at
once and each term reduces to $\| q_d \| \| k_d \| \cos \alpha_d$. For
$\Delta \neq 0$ the phase grows fastest in the early, high-frequency chunks
of Eq.~\eqref{eq:rope_freq}, whose contributions therefore fall away first.
The coordinate-aligned source token is thus favoured over displaced tokens
of comparable content similarity, and attention resolves by coordinate
rather than by correspondence.

In our setting the source $I$ and the composite $I_g$ depict the same object
in the same scene, so the coordinate-aligned source token for an edited part
is that part's pre-edit texture, the one location from which appearance must
not be copied. The part is therefore repainted at its pre-edit coordinates,
as visualized in Fig.~3a of the main paper. 
This localizes the problem to the high-frequency chunks. Accordingly,
$\mathcal{M}_{low}(S_{lf})$ in Eq.~(12) of the main paper assigns to both
channels of pair $d$ the scale $s_d$ of Eq.~\eqref{eq:ramp}, so that the elementwise
product attenuates the high-frequency chunks and amplifies the low-frequency
ones.

\section{CS-ARAP Algorithm}
\label{sec:algorithm}

Algorithm~\ref{alg:smooth-arap} provides the full pseudocode for
our CS-ARAP deformation. The key modification over the standard
ARAP occurs in the local step: rather than
computing per-vertex rotations independently via SVD of the raw
covariance $S_i$, we first smooth $S_i$ by taking a convex
combination with the mean covariance of its 1-ring neighborhood. This acts as a low-pass filter on the affine
transformations encoded in the covariances, suppressing the
high-frequency rotation spikes that arise under sparse handle
configurations. The global step replaces the standard Laplacian
solve with a bi-Laplacian system that penalizes changes in
discrete curvature, further reducing positional kinks at the constraint
boundaries. In Eq.5 of the main paper, ‘fixed’ refers to the prescribed handle constraints; non-handle vertices remain free variables and are solved by the CS-ARAP global step.

\begin{algorithm}[]
\caption{CS-ARAP}
\label{alg:smooth-arap}
\KwIn{Rest mesh $(V, F)$, constraint indices $\mathcal{B}$,
constraint positions $\hat{V}_\mathcal{B}$, smoothing parameters
$\lambda$, $\beta$, convergence threshold $\epsilon$}
\KwOut{Deformed mesh $V'$}
\tcp{Precomputation}
$L \leftarrow \text{CotangentLaplacian}(V, F)$ \;
$M \leftarrow \text{MassMatrix}(V, F)$ \;
$A \leftarrow (1 - \lambda) L + \lambda L M^{-1} L$ \;
Factor $A$ with constrained Cholesky (fixing rows/cols in
$\mathcal{B}$) \;
$V' \leftarrow V$ \;
\tcp{Alternating Minimization}
\Repeat{$\|V' - V'_{\text{old}}\| / \|V\| < \epsilon$}{
    $V'_{\text{old}} \leftarrow V'$ \;
    \tcp{Local Step: Smoothed Rotation Update}
    \For{each vertex $i$}{
        $S_i \leftarrow \sum_{j} w_{ij}
        (p_i - p_j)(p'_i - p'_j)^T$
    }
    \For{each vertex $i$}{
        $\tilde{S}_i \leftarrow (1 - \beta)\, S_i +
        \beta \left( \frac{1}{|\mathcal{N}_i|}
        \sum_{j \in \mathcal{N}_i} S_j \right)$
        \tcp*{Covariance smoothing}
    }
    \For{each vertex $i$}{
        $U \Sigma V^T \leftarrow \text{SVD}(\tilde{S}_i)$ \;
        $\tilde{R}_i \leftarrow U V^T$
        \tcp*{Project to $SO(3)$}
        \lIf{$\det(\tilde{R}_i) < 0$}{
            flip smallest singular value}
    }
    \tcp{Global Step: Bi-Laplacian Position Solve}
    \For{each vertex $i$}{
        $b_i \leftarrow (1 - \lambda) \sum_{j} w_{ij}\,
        \tilde{R}_i\, (p_i - p_j) +
        \lambda\, (L M^{-1} \tilde{R}(LV))_i$
    }
    Solve $A\, V' = b$ subject to
    $V'_\mathcal{B} = \hat{V}_\mathcal{B}$ \;
}
\Return $V'$
\end{algorithm}

\begin{figure*}[!htbp]
\centering
\includegraphics[width=\linewidth]{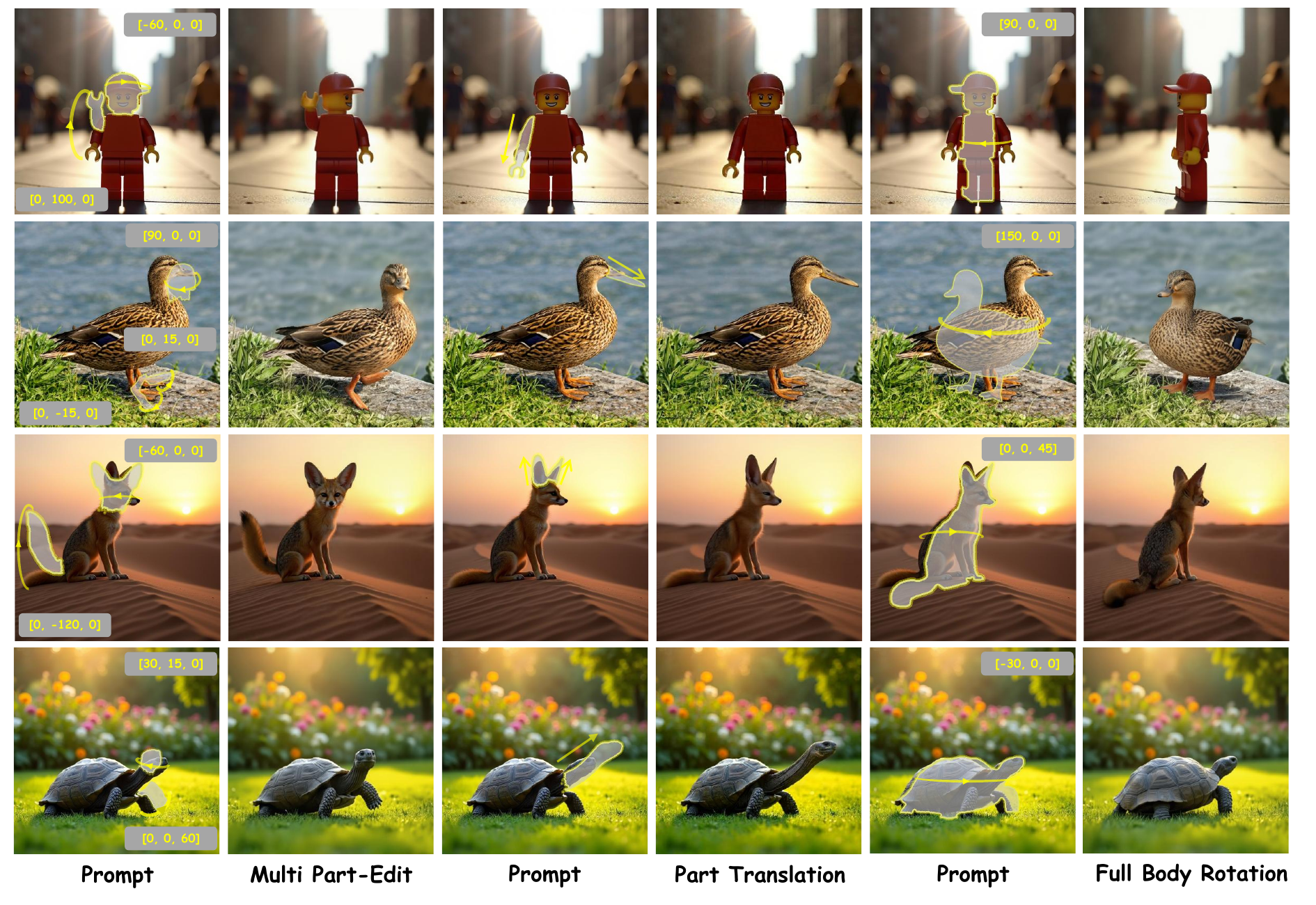}
\caption{Additional results across editing operations:
multi-part editing, part translation, and full-body rotation.}
\label{fig:fig6_extended}
\end{figure*}
\section{Additional Ablations}
\label{sec:ablations}

\subsection{Effect of IP-Adapter Conditioning on SDS}
\label{sec:ipadapter_ablation}

A natural question is whether IP-Adapter conditioning is
necessary for SDS refinement, or whether vanilla SDS gradients
alone are sufficient to improve global shape quality.
Table~\ref{tab:ipadapter} answers this directly. Without
IP-Adapter, SDS lacks a visual identity anchor and pushes the
geometry toward generic plausible shapes that may not match
the source object. The drop in CLIP-I confirms that vanilla
SDS loses object identity, while SDS conditioned via IP-Adapter
preserves it throughout refinement.

\subsection{Appearance Transfer Module Comparison}
\label{sec:appearance_ablation}

Table~\ref{tab:appearance_ablation} compares our full appearance
transfer module against two closely related baselines that
use cross-image attention. MasaCtrl was
designed for consistent multi-concept generation rather than
geometry-aware editing, and lacks the structure-aware guidance
needed to prevent structural collapse when the source and target
geometries differ significantly. Cross-Image-Attn
follows a similar trend, leaving foreground colors and seams misaligned
with the source. The qualitative differences are visible in
Fig.~\ref{fig:tex_trans_methods}, where both baselines introduce
color bleeding and structural artifacts that our module
avoids.

\subsection{3D Lifting Model Comparison}
\label{sec:lifting}

SAM3D is chosen as our lifting backbone based on both efficiency
and reconstruction quality. Table~\ref{tab:lifting_comparison}
compares it against three recent alternatives. Despite being the
fastest by a wide margin, SAM3D achieves the best semantic
alignment scores under both ULIP and Uni3D, making it the natural
choice.
Other methods score lower at 600s and 200s, respectively, making them impractical for
interactive editing.

\subsection{Part Localization Ablation}
\label{sec:bbox}

Accurate part localization is a prerequisite for correct
part segmentation. An imprecise bounding box causes the
downstream segmentation to either miss part of the target
part or include background regions. Fig.~\ref{fig:bbox_ablation}
compares RADI3N's localization against YoloWorld and OwlV2 across
a diverse set of objects. Both baselines struggle with objects
that have irregular boundaries or low contrast against the
background, producing loose boxes that propagate errors into
part segmentation. RADI3N's Qwen-guided approach consistently
produces tighter, more semantically grounded localizations,
directly enabling more accurate part-level editing.

\subsection{Part Segmentation}
\label{sec:part_seg}
\begin{figure}[t]
    \centering
    \includegraphics[width=\linewidth]{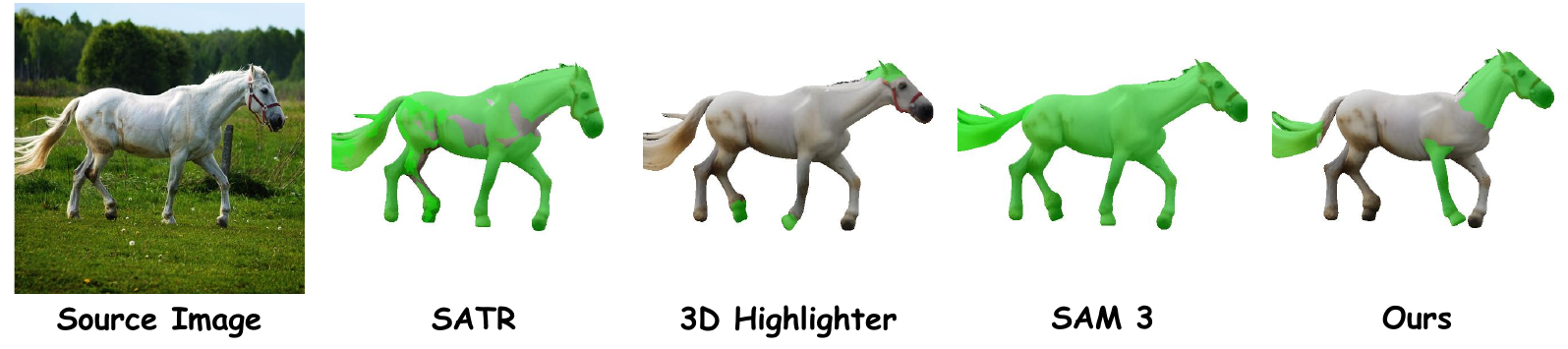}
    \caption{\textbf{Part Segmentation Comparison.}}
    \label{fig:ablation_seg}
\end{figure}

\begin{figure*}[!htbp]
\centering
\includegraphics[width=0.5\textwidth]{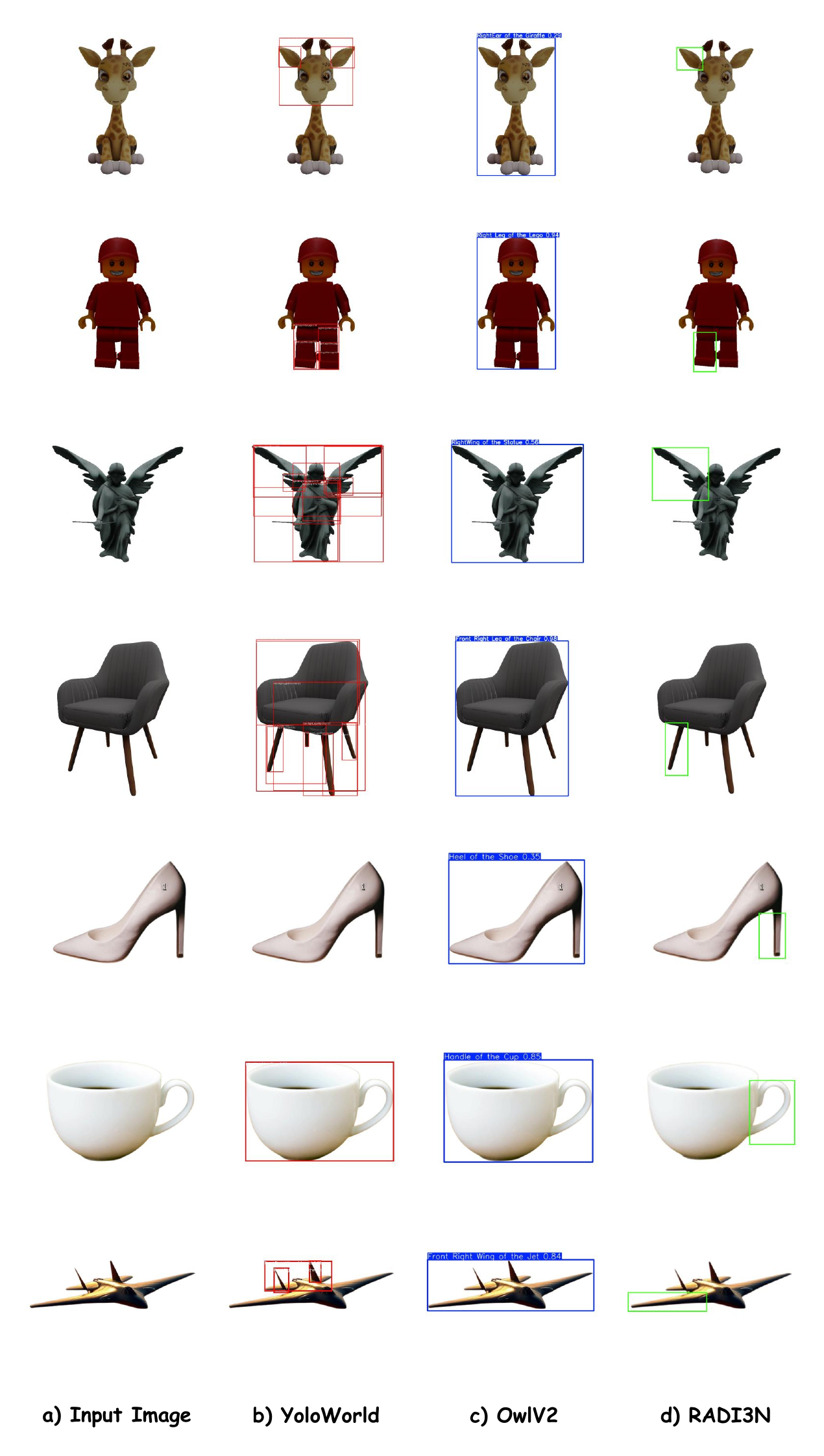}
\caption{Bounding box detection comparison across YoloWorld,
OwlV2, and RADI3N. Tighter localizations lead directly to
more accurate part segmentation in downstream stages.}
\label{fig:bbox_ablation}
\end{figure*}

Against SATR, 3D-Highlighter,
and SAM~3 on a horse (head, front-right leg, tail;
Fig.~\ref{fig:ablation_seg}): SATR masks are overly broad, 3D-Highlighter
fails on all three parts, and so does SAM~3 which also cannot disambiguate repeated
structures such as specific limbs. RADI3N yields anatomically coherent masks on
all three parts. Across 25 objects with multiple target parts, GPT-5.6 (high
reasoning) preferred our segmentation in 24/25 cases.

\section{Evaluation Protocol}
\label{sec:eval_protocol}

\subsection{Part Angular Accuracy}
\label{sec:hunyuan_eval}

Angular accuracy in the main paper is measured by lifting edited
images back to 3D via SAM3D, the same model used in our
pipeline. To confirm this does not introduce circular evaluation
bias, we re-run the full angular evaluation using
Hunyuan3D as an independent lifting method.
The aim is to confirm that the evaluator is independent of the editing
pipeline. As shown in Table~\ref{tab:synthetic_eval}, RADI3N
maintains the best angular accuracy under Hunyuan3D evaluation,
with numbers slightly worse than the SAM3D-based metric due to
differences in reconstruction between the two models. The
consistent ranking across both evaluators confirms that our
reported angular accuracy is not an artifact of metric circularity.

\begin{table}[!t]
\small
\centering
\caption{Part angular accuracy evaluated using
Hunyuan3D on GeoBench for 3D reconstruction, independent
of the SAM3D used in the editing pipeline.}
\label{tab:synthetic_eval}
\begin{tabular}{l cc}
\toprule
{\textbf{Method}} & MAAE$^\circ$ $\downarrow$
& Acc@5$^\circ$ $\uparrow$ \\
\midrule
Image Sculpting & 16.61 & 0.354 \\
FlowDrag       & 23.93 & 0.141 \\
Nano Banana 2   & 11.27 & 0.479 \\
\midrule
\rowcolor[gray]{0.9}
\textbf{RADI3N}                        & \textbf{8.41}
& \textbf{0.746} \\
\bottomrule
\end{tabular}
\end{table}

\subsection{Full-Body Rotation Evaluation}
\label{sec:fullbody}

For full-body rotations, part-level angular evaluation is
not applicable since the entire object moves. We instead
measure object orientation using Orient
Anything, which estimates
whole-object pose directly from the edited image.
Table~\ref{tab:fullbody_rotation} reports results on this
subset alongside standard geometric fidelity metrics. RADI3N
achieves the highest OA-Acc@5° while also maintaining the
lowest WE and D-RMSE, indicating that our 3D mesh proxy
faithfully captures the intended full-body transformation.

\begin{table}[!t]
\small
\centering
\caption{Full-body rotation evaluation. OA-Acc@5$^\circ$ (whether the desired angle is within the 5-degree range of the angle)is
measured via Orient Anything.}
\label{tab:fullbody_rotation}
\begin{tabular}{l ccc}
\toprule
{\textbf{Method}} & WE $\downarrow$ & D-RMSE $\downarrow$
& OA-Acc@5$^\circ$ $\uparrow$ \\
\midrule
FreeFine      & 0.126 & 3.11 & 0.573 \\
Object3DiT        & 0.117 & 3.38 & 0.741 \\
Qwen-Img-Edit & 0.096 & 2.80 & 0.504 \\
\midrule
\rowcolor[gray]{0.9}
\textbf{RADI3N} & \textbf{0.073} & \textbf{2.12} & \textbf{0.764} \\
\bottomrule
\end{tabular}
\end{table}

\subsection{Part Translation Evaluation}
\label{sec:translation_eval}

Part translations do not involve rotation and are therefore
not captured by angular metrics.
Table~\ref{tab:translation_eval} shows that RADI3N achieves
better subject consistency (SUBC) and background consistency
(BC) than GoodDrag on DragBench, reflecting the benefit of
our explicit 3D mesh proxy for maintaining spatial coherence
during translation edits.

\begin{table}[!t]
\small
\centering
\caption{Part-translation consistency evaluation.}
\label{tab:translation_eval}
\begin{tabular}{l ccc}
\toprule
{\textbf{Method}} & SUBC $\uparrow$ & DINOv2 $\downarrow$
& BC $\uparrow$ \\
\midrule
GoodDrag & 0.804 & 1888.41 & 0.923 \\
\midrule
\rowcolor[gray]{0.9}
\textbf{RADI3N} & \textbf{0.826} & \textbf{1863.17}
& \textbf{0.941} \\
\bottomrule
\end{tabular}
\end{table}




\subsection{Sensitivity Analysis}
\label{sec:sensitivity}

\paragraph{Choice of $\lambda$ and $\beta$.}
The bi-Laplacian weight $\lambda$ controls the trade-off between
local rigidity ($E_{\text{ARAP}}$) and position smoothness
($E_{\text{blp}}$). We set $\lambda = 0.95$ to heavily favor
$C^2$ positional smoothness, as our primary goal is to produce
a spike-free initialization for the subsequent SDS refinement.
Lower values of $\lambda$ preserve local rigidity more faithfully
but reintroduce positional kinks at constraint boundaries. The
covariance smoothing parameter $\beta \in [0.0, 0.2]$ controls
the degree of rotation field smoothing. We find that $\beta = 0$
(no smoothing) suffices for small edits where the standard ARAP
rotation field is already well-behaved, while $\beta = 0.2$ is
needed for large articulations where neighboring rotations
diverge significantly. In practice, we set $\beta$ proportional
to the edit magnitude: larger handle displacements use a higher $\beta$.

\noindent\textbf{Stage-wise error propagation.}
To quantify how much each stage of the pipeline governs final output quality,
we run a stage-wise sensitivity analysis over the three stages: 3D lifting,
mesh deformation, and appearance transfer. For each stage we split the
evaluation set into bottom- and top-$50$-percentile buckets according to that
stage's own quality metric (Uni3D, D-RMSE, and FID,
respectively) and measure the resulting difference in final output quality
between the two buckets. Final quality differs by $8.73\%$ across the
3D-lifting buckets, but by only $1.78\%$ (mesh deformation) and $3.55\%$
(appearance transfer) across the two stages we contribute. Output quality is
therefore governed primarily by the upstream reconstructor rather than by our
components, which remain stable under substantial variation in their inputs.
This is consistent with the failure modes in Sec.~\ref{sec:failure}, where poor
SAM3D geometry propagates into degraded edits
(Figs.~\ref{fig:mesh_failures}--\ref{fig:failures}), and with
Tab.~\ref{tab:synthetic_eval}, where substituting an independent lifter preserves the
method ranking.

\noindent\textbf{Number of SDS iterations.}
Refinement quality is stable across iteration counts. Reducing
$N_{\text{iter}}$ to $250$, one quarter of our default, yields
WE $0.075$ / D-RMSE $1.99$, close to the $1{,}000$-iteration result
($0.067$ / $1.91$) and still ahead of all baselines in
Tab.2 of main paper. Refinement therefore converges well before the
iteration budget expires, and $N_{\text{iter}} = 1{,}000$ is a conservative
quality-maximizing choice rather than a brittle setting. Lower counts remain a
practical option when interactive latency matters.

\noindent\textbf{Low-frequency amplification $S_{lf}$.} The strength
$S_{lf}$ controls how strongly the retained low-frequency chunks are
amplified. Quality improves
monotonically up to $S_{lf} = 2$ and then plateaus. We therefore
fix $S_{lf} = 2$ throughout. Very low values leave insufficient attention on
the source and the transferred appearance is washed out, while beyond the
plateau the additional gain has no effect.

\subsection{Runtime Analysis of Individual Components}
\label{sec:runtime}
A single edit takes 95-100s end-to-end on an NVIDIA H100 GPU. 3D lifting via
SAM3D accounts for around 25s. The CS-ARAP deformation (Stage 1) converges in
less than 2s on an average. SDS refinement (Stage 2) dominates remaining runtime at
45-50s, as each iteration backpropagates through the diffusion
prior. Appearance blending adds a
further 20-30s.

\section{DragBench Results}
\label{sec:dragbench}
Table~\ref{tab:dragbench} reports results on the full DragBench
benchmark. On DragBench, RADI3N achieves the best geometric fidelity
(WE 0.074, D-RMSE 2.12) and angular accuracy (MAAE $7.9^\circ$,
Acc@5$^\circ$ 69.8\%). It reduces MAAE by 52\% over the strongest
baseline, Image Sculpting ($16.5^\circ$), while more than $1.6\times$
its Acc@5$^\circ$ (42.4\%). Nano Banana 2 achieves the best CLIP-I
(0.918), FID (161.4), and DINOv2 distance (1893.5), followed closely
by Qwen-Img-Edit (0.908, 162.9) and RADI3N (0.905, 164.3, 1904.1),
which remains competitive on appearance and identity preservation.
RADI3N achieves the best SUBC (0.808), and Qwen-Img-Edit has the
highest BC (0.929), with RADI3N close behind (0.925), indicating
strong background preservation. The larger absolute MAAE values
across nearly all methods reflect the more diverse and challenging edit
types present in DragBench compared to GeoBench.

\section{User Study}
\label{sec:user_study}

We conducted a perceptual user study to evaluate whether the
geometric improvements reflected in our quantitative metrics
translate to human preference. Participants were shown a source
image alongside the editing prompt and asked to select the output
that best reflects the intended geometric and angular transformation
while remaining visually realistic and confined to the masked
region. Critically, participants were instructed to pay close
attention to texture consistency outside the masked area, ensuring
that background preservation was weighted alongside geometric
fidelity in their judgments. They were shown a prompt image similar to our qualitative comparison diagrams and were also given the edit instruction.

The study included 34 participants evaluating randomly ordered
outputs from RADI3N and six baselines: FreeFine, FlowDrag,
GoodDrag, Qwen-Image-Edit, Nano Banana 2, and ImageSculpting. RADI3N was preferred in \textbf{79.6\%}
of comparisons, with no single baseline exceeding 12\%. The margin
is notably large; the closest competitor (FlowDrag) accounts
for only 12\% of preferences, suggesting that RADI3N consistently outperforms other methods in both parametric angular edits and texture alignment with the input.

\begin{figure*}[!htbp]
\centering
\includegraphics[width=\linewidth]{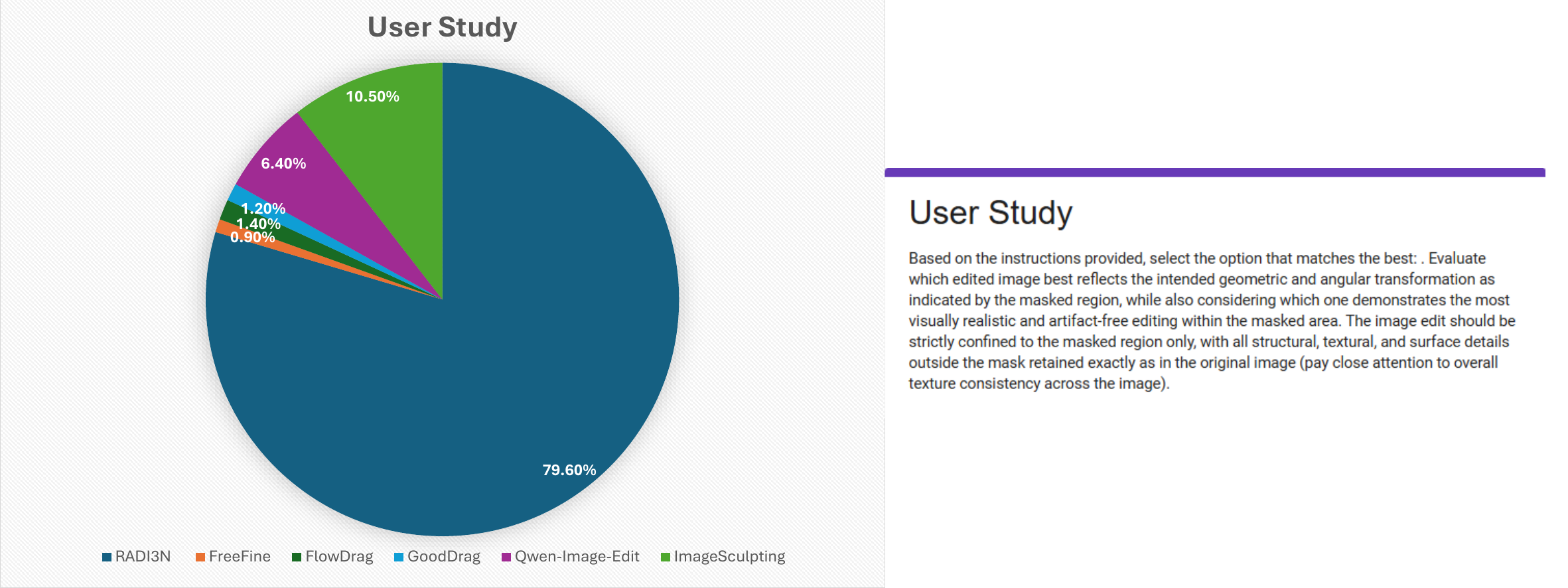}
\caption{User Study: Comparing our method with FreeFine, GoodDrag, FlowDrag, Qwen-Image-Edit, ImageSculpting}
\label{fig:user_study}
\end{figure*}

\section{Additional Qualitative Results}
\label{sec:qual}

\subsection{Extended Method Comparisons}
\label{sec:qual_comparison}

Fig.~\ref{fig:fig5_extended} extends the qualitative comparison
from Fig.4 of the main paper to additional object categories.
Across all examples, drag-based baselines struggle with
large articulations, producing warping artifacts, inconsistent texture, and poor adherence to prompts. RADI3N consistently
preserves the surrounding geometry while executing the intended
transformation, reflecting the benefit of an explicit 3D proxy
over purely 2D deformation approaches.


\begin{table*}[!t]
\small
\centering
\caption{Quantitative comparison on \textbf{DragBench}.}
\label{tab:dragbench}
\resizebox{\textwidth}{!}{
\begin{tabular}{l cccc cc ccc}
\toprule
{\textbf{Method}} & \multicolumn{4}{c}{\textbf{Geometric Fidelity}}
& \multicolumn{2}{c}{\textbf{Appearance \& Identity}}
& \multicolumn{3}{c}{\textbf{Consistency}} \\
\cmidrule(lr){2-5} \cmidrule(lr){6-7} \cmidrule(lr){8-10}
& WE $\downarrow$ & D-RMSE $\downarrow$ & MAAE$^{\circ}$ $\downarrow$
& Acc@5$^{\circ}$ $\uparrow$ & CLIP-I $\uparrow$ & FID $\downarrow$
& SUBC $\uparrow$ & DINOv2 $\downarrow$ & BC $\uparrow$ \\
\midrule
GoodDrag
    & 0.094 & 3.06 & 29.8 & 0.074
    & 0.877 & 184.3 & 0.774 & 1933.4 & 0.873 \\
FlowDrag
    & 0.086 & 2.37 & 23.7 & 0.156
    & 0.884 & 165.8 & 0.783 & 1923.1 & 0.910 \\
FreeFine
    & 0.125 & 2.76 & 22.5 & 0.252
    & 0.879 & 168.2 & 0.789 & 1922.9 & 0.798 \\
Image Sculpting
    & 0.091 & 2.80 & \underline{16.5} & \underline{0.424}
    & 0.881 & 176.4 & 0.783 & 1929.4 & 0.881 \\
Qwen-Img-Edit
    & \underline{0.084} & 2.47 & 20.5 & 0.213
    & \underline{0.908} & \underline{162.9}
    & \underline{0.799} & 1918.2 & \textbf{0.929} \\
Nano Banana 2
    & 0.089 & \underline{2.30} & 18.7 & 0.396
    & \textbf{0.918} & \textbf{161.4}
    & 0.790 & \textbf{1893.5} & 0.911 \\
\midrule
\rowcolor[gray]{0.9}
\textbf{RADI3N}
    & \textbf{0.074} & \textbf{2.12} & \textbf{7.9} & \textbf{0.698}
    & 0.905 & 164.3
    & \textbf{0.808} & \underline{1904.1} & \underline{0.925} \\
\bottomrule
\end{tabular}
}
\end{table*}

\subsection{Additional Editing Operations}
\label{sec:qual_operations}

Fig.~\ref{fig:fig6_extended} shows additional results for
multi-part editing, part translation, and full-body rotation,
complementing the examples in Fig.5 of the main paper.
Multi-part edits are particularly challenging as each part
must deform consistently while maintaining global structural
plausibility. The 3D mesh segmentation in RADI3N handles this
naturally, since each part operates in a shared 3D coordinate
frame. RADI3N's ability to handle large part-based and full-body deformations parametrically is not feasible for comparative methods.


\subsection{3D Part Mask Propagation}
\label{sec:qual_mask}

\begin{figure*}[!htbp]
\centering
\includegraphics[width=\linewidth]{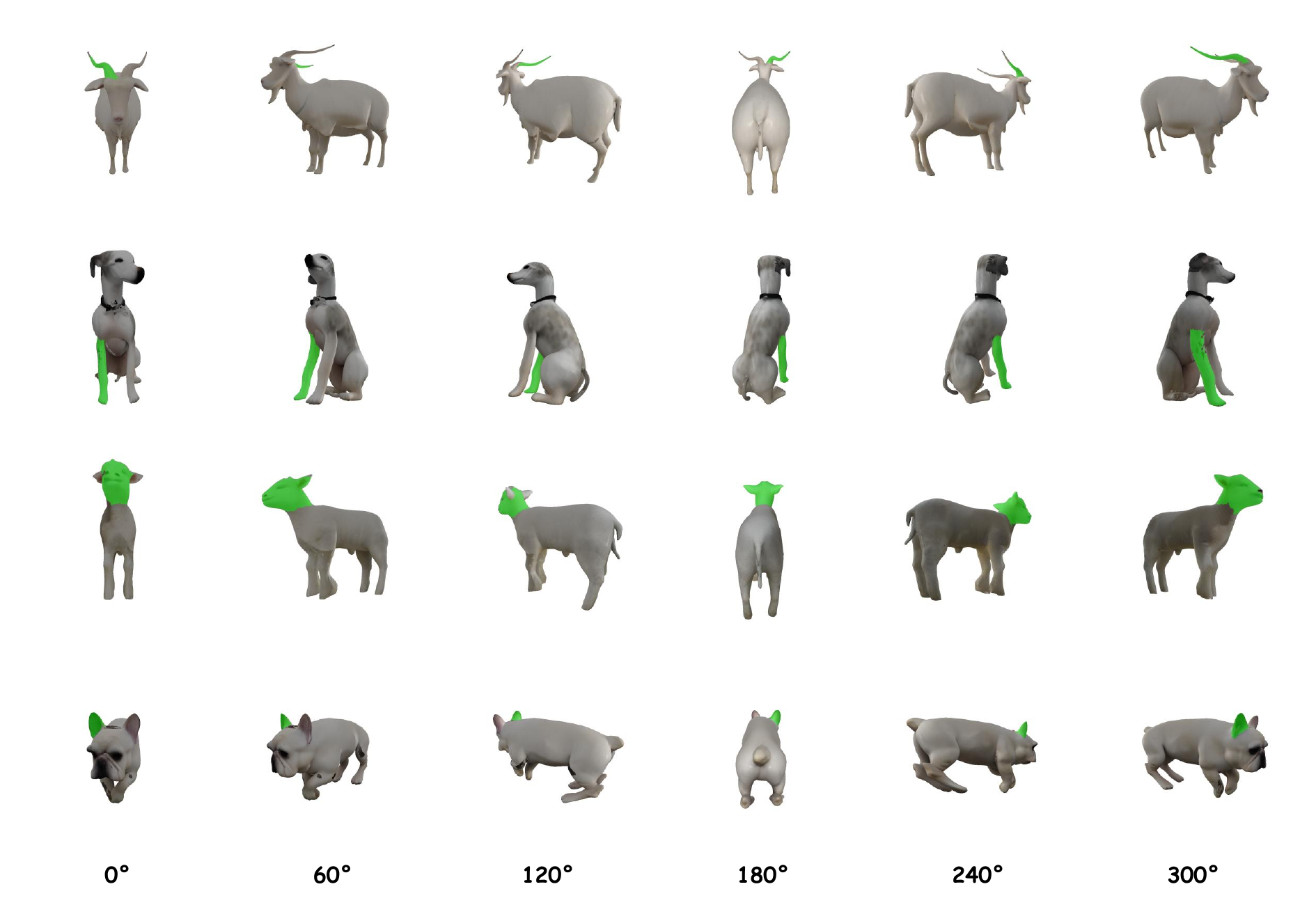}
\caption{3D mask propagation via UV-space majority voting.
Per-view 2D segmentations are lifted to consistent 3D part
labels on the mesh surface.}
\label{fig:fig3d_prop}
\end{figure*}

\begin{figure*}[!htbp]
\centering

\includegraphics[width=0.7\linewidth]{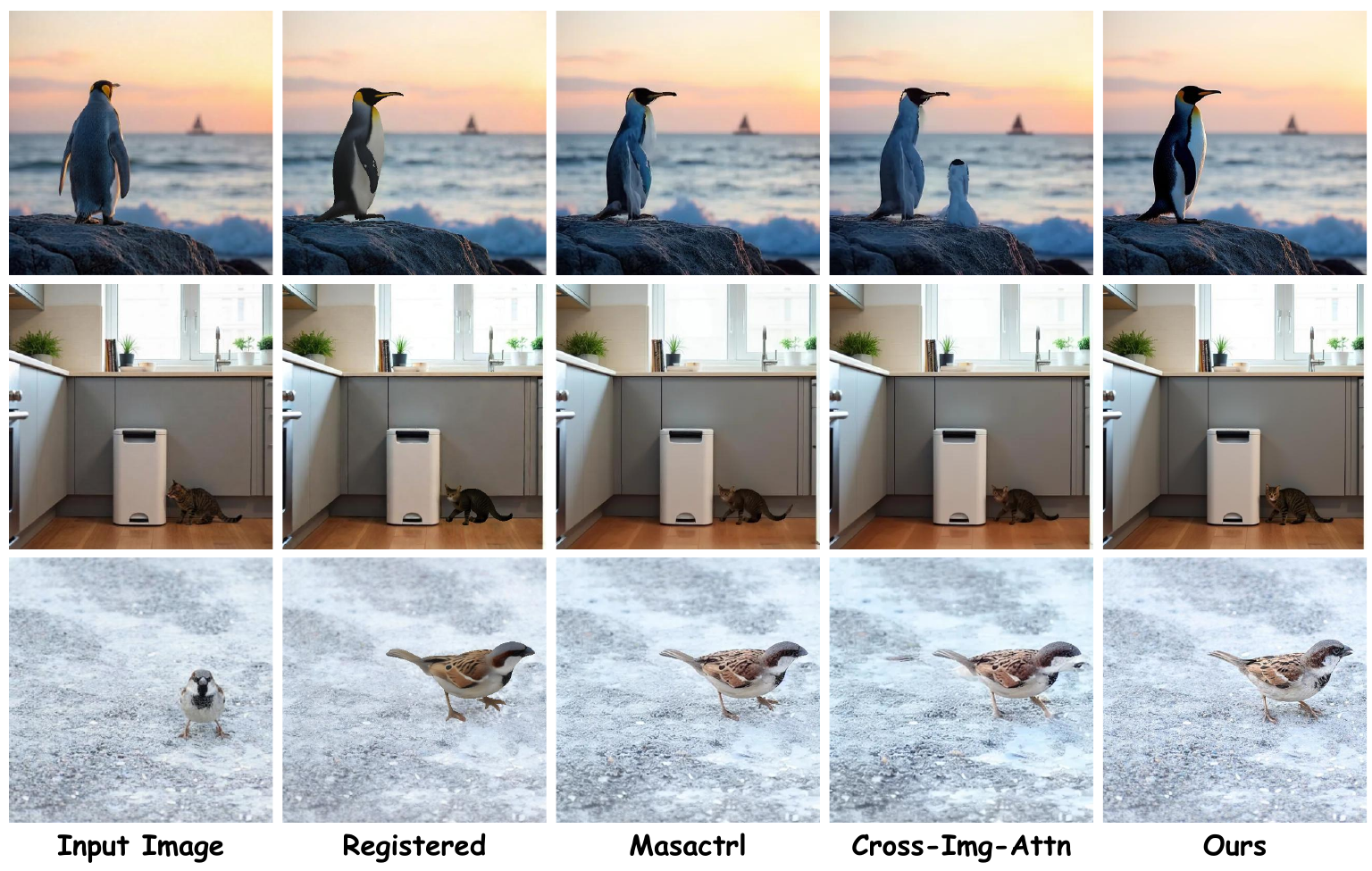}

\caption{Qualitative comparison of appearance transfer methods.
MasaCtrl and Cross-Image-Attn introduce color bleeding and
structural artifacts that the full RADI3N module avoids.}
\label{fig:tex_trans_methods}

\end{figure*}

Fig.~\ref{fig:fig3d_prop} illustrates the UV-space multi-view majority voting used for 3D part segmentation. Per-view 2D
segmentations are projected into a shared UV atlas and
aggregated, producing view-consistent part labels that transfer
cleanly to the 3D mesh surface. This approach handles parts
that are partially occluded from any single viewpoint by
leveraging information from complementary camera angles.

\begin{table}[!t]
\footnotesize

\centering
\caption{Effect of IP-Adapter conditioning on SDS refinement.}
\label{tab:ipadapter}
\begin{tabular}{l cccc}
\toprule
{\textbf{Configuration}} & WE $\downarrow$ & D-RMSE $\downarrow$
& CLIP-I $\uparrow$ & FID $\downarrow$ \\
\midrule
Vanilla SDS              & 0.146 & 2.12 & 0.898 & 152.25 \\
\midrule
\rowcolor[gray]{0.9}
\textbf{SDS + IP-Adapter (ours)} & \textbf{0.067} & \textbf{1.91}
& \textbf{0.924} & \textbf{148.47} \\
\bottomrule
\end{tabular}
\end{table}



\begin{table}[!t]
\small
\centering
\caption{Appearance transfer module comparison on GeoBench.}
\label{tab:appearance_ablation}
\begin{tabular}{l cc}
\toprule
{\textbf{Method}} & CLIP-I $\uparrow$ & FID $\downarrow$ \\
\midrule
MasaCtrl             & 0.681 & 191.34 \\
Cross-Image-Attn & 0.803 & 199.82 \\
\midrule
\rowcolor[gray]{0.9}
\textbf{RADI3N}                              & \textbf{0.924} & \textbf{148.47} \\
\bottomrule
\end{tabular}
\end{table}

\begin{table}[!t]
\small
\centering
\caption{\textbf{3D Lifting Comparison.} ULIP and
Uni3D scores measure perceptual similarity between
reconstructed geometry and the source image.}
\label{tab:lifting_comparison}
\begin{tabular}{l c cc}
\toprule
{\textbf{Model}} & \textbf{Time (s)} $\downarrow$
& ULIP $\uparrow$ & Uni3D $\uparrow$ \\
\midrule
SV3D       & 600 & 0.1124 & 0.3362 \\
HY3D-2.1 & 200 & 0.1293 & 0.3546 \\
TRELLIS            &  45 & 0.1473 & 0.3698 \\
\midrule
\rowcolor[gray]{0.9}
\textbf{SAM3D} & \textbf{25} & \textbf{0.1488}
& \textbf{0.3707} \\
\bottomrule
\end{tabular}
\end{table}

\subsection{Scene Registration Ablation}
\label{sec:qual_registration}

\begin{figure*}[!htbp]
\centering
\includegraphics[width=\linewidth]{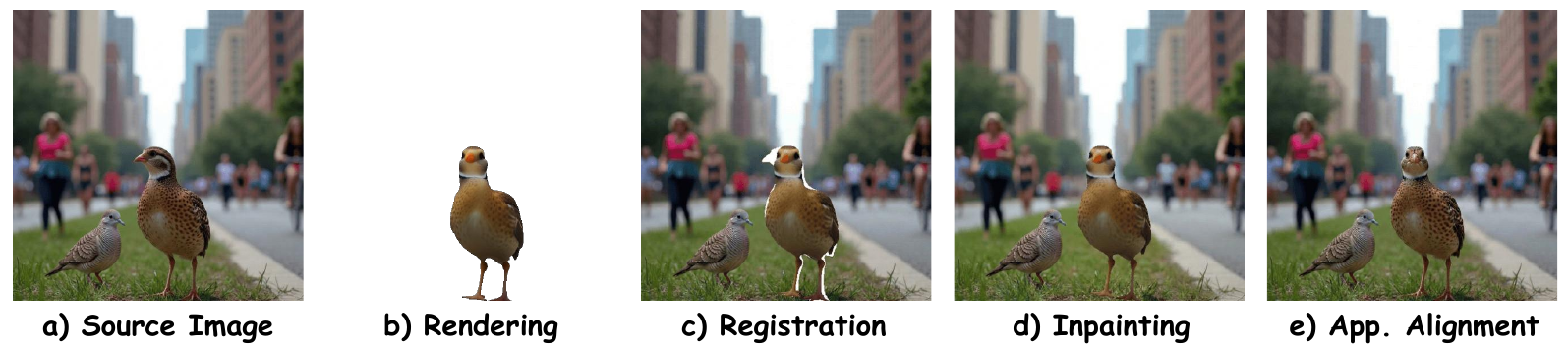}
\caption{Scene registration ablation. Each column removes one
component of the blending pipeline, showing its contribution
to the final composite quality.}
\label{fig:regi_ablation}
\end{figure*}

Fig.~\ref{fig:regi_ablation} ablates the scene registration
and blending components. Without background inpainting, the
original object leaves a visible hole in the scene after
removal. Without the scale and position correction,
minor discrepancies between the rendered mesh and the input
image produce misaligned composites. Both steps are necessary
for clean foreground-background integration, particularly for
edits involving significant depth changes such as z-translations
where the object scale in the image changes noticeably.

\section{Failure Cases and Limitations}
\label{sec:failure}

\begin{figure*}[!htbp]
\centering
\includegraphics[width=\linewidth]{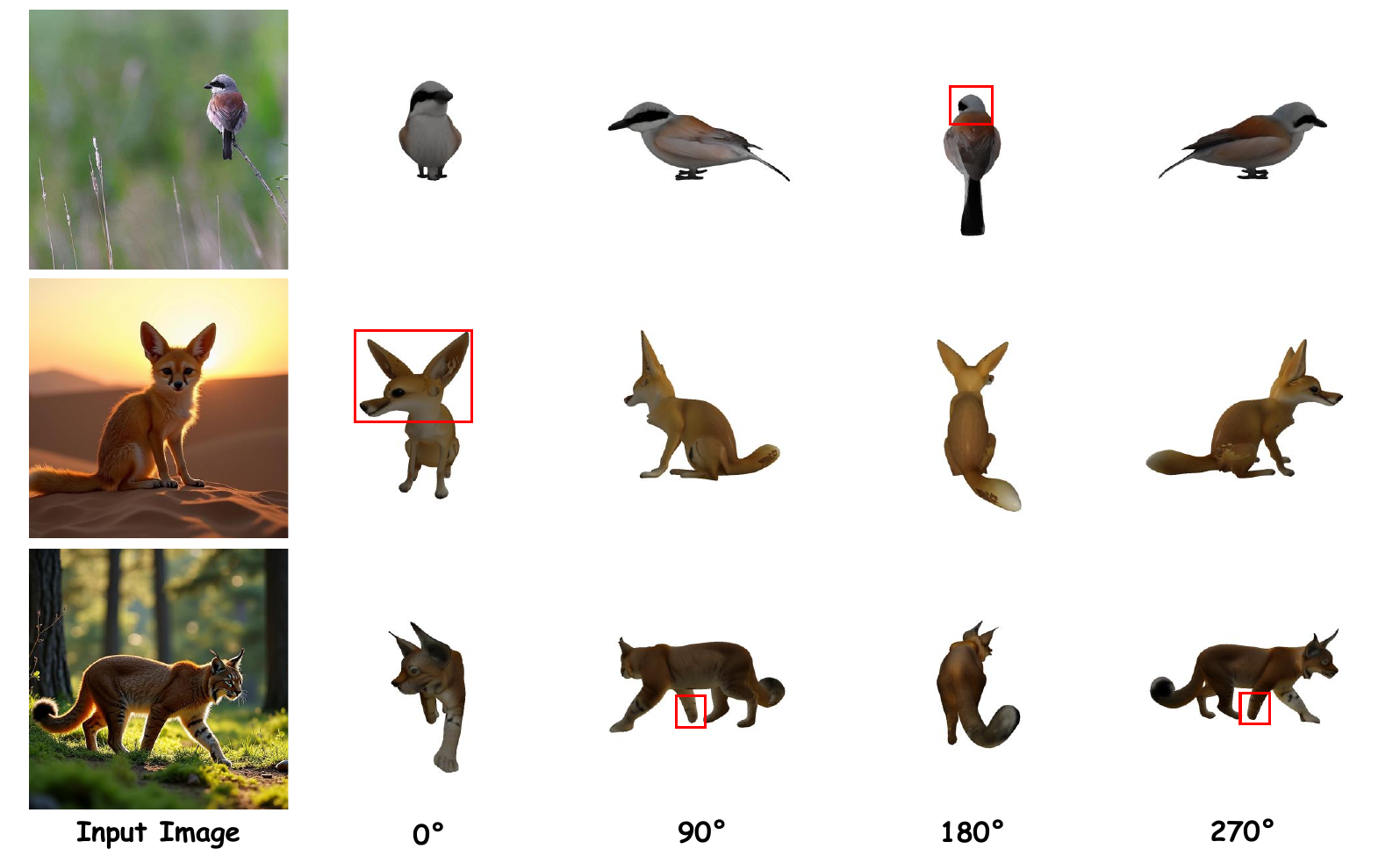}
\caption{Failure cases in mesh generation. Thin structures and
heavily occluded surfaces produce incomplete or noisy meshes
that limits edit quality.}
\label{fig:mesh_failures}
\end{figure*}

Fig.~\ref{fig:mesh_failures} shows cases where the 3D mesh
reconstruction from SAM3D produces poor geometry, typically
for objects with thin structures, small sizes, high surface complexity, or
severe self-occlusion in the input view. Since our entire
pipeline builds on this proxy mesh, reconstruction errors at
this stage propagate and limit the quality of downstream edits.
Improving robustness to single-view reconstruction failure is
the most impactful direction for future work.

\begin{figure*}[!htbp]
\centering
\begin{minipage}[c]{0.56\textwidth}
    \centering
    \includegraphics[width=\linewidth]{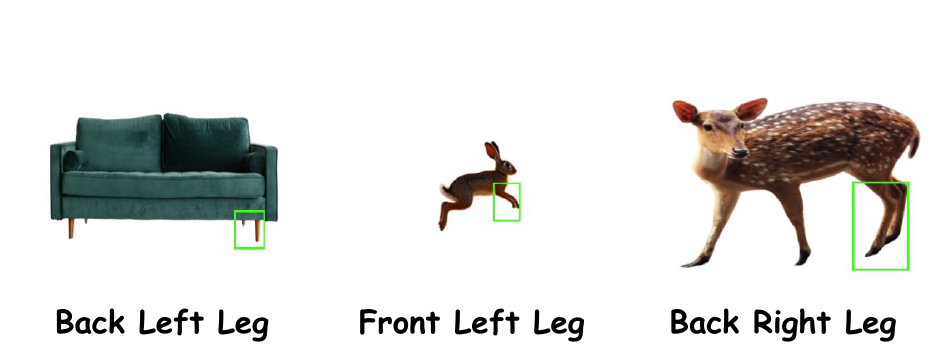}
\end{minipage}\hfill
\begin{minipage}[c]{0.40\textwidth}
    \centering
    \includegraphics[width=\linewidth]{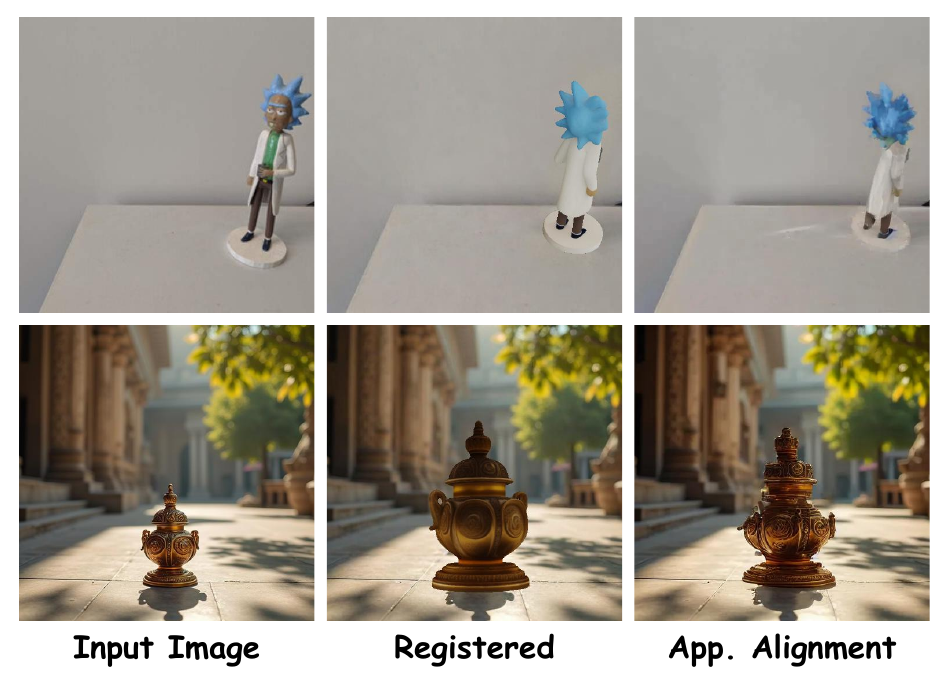}
\end{minipage}
\par\vspace{4pt}
\begin{minipage}[t]{0.56\textwidth}
    \caption{Part selection failures. Clustered or similar
    structures confuse the VLM, leading to merged or mislocalized
    bounding boxes.}
    \label{fig:failures_bbox}
\end{minipage}\hfill
\begin{minipage}[t]{0.40\textwidth}
    \caption{Appearance transfer failures under large rotations
    and extreme Z-translations, where hidden surfaces have no
    source texture reference.}
    \label{fig:failures}
\end{minipage}
\end{figure*}

Part selection failures are shown in
Fig.~\ref{fig:failures_bbox}. When object parts are spatially
clustered or visually similar, the VLM occasionally produces
bounding boxes that merge adjacent parts or select the wrong instance entirely. This is most pronounced for repeated
structures such as animal legs or symmetric objects, where
left-right disambiguation is ambiguous from natural language
alone and can be part of a future exploration.

Fig.~\ref{fig:failures} shows appearance transfer failures.
Large full-body rotations and extreme z-translations expose
previously hidden surfaces with no source texture information,
causing the appearance transfer to hallucinate inconsistent
texture in those regions. This is a fundamental limitation of
single-view input rather than a failure of the transfer
mechanism itself.

\section{VLM Parsing and Part Selection Examples}
\label{sec:vlm}

Fig.~\ref{fig:prompts_part_sel} and
Fig.~\ref{fig:prompts_obj_sel} show representative examples
of the two-stage VLM parsing used in RADI3N. The object
selection stage (Fig.~\ref{fig:prompts_obj_sel}) described in the main paper's Object Selection paragraph decomposes a
natural language prompt into a localization target and a list
of associated part names. The part selection stage (Fig.~\ref{fig:prompts_part_sel}) then maps
these part names to 3D mesh regions via the UV-space
segmentation pipeline. Across diverse prompts, the VLM
correctly identifies semantically meaningful part boundaries
even for complex multi-word descriptions, though it
occasionally fails on spatially ambiguous references such as
``left'' vs ``right'' limbs when the object orientation is
not unambiguous from the image.

\begin{strip}
\centering
\includegraphics[width=0.8\textwidth]{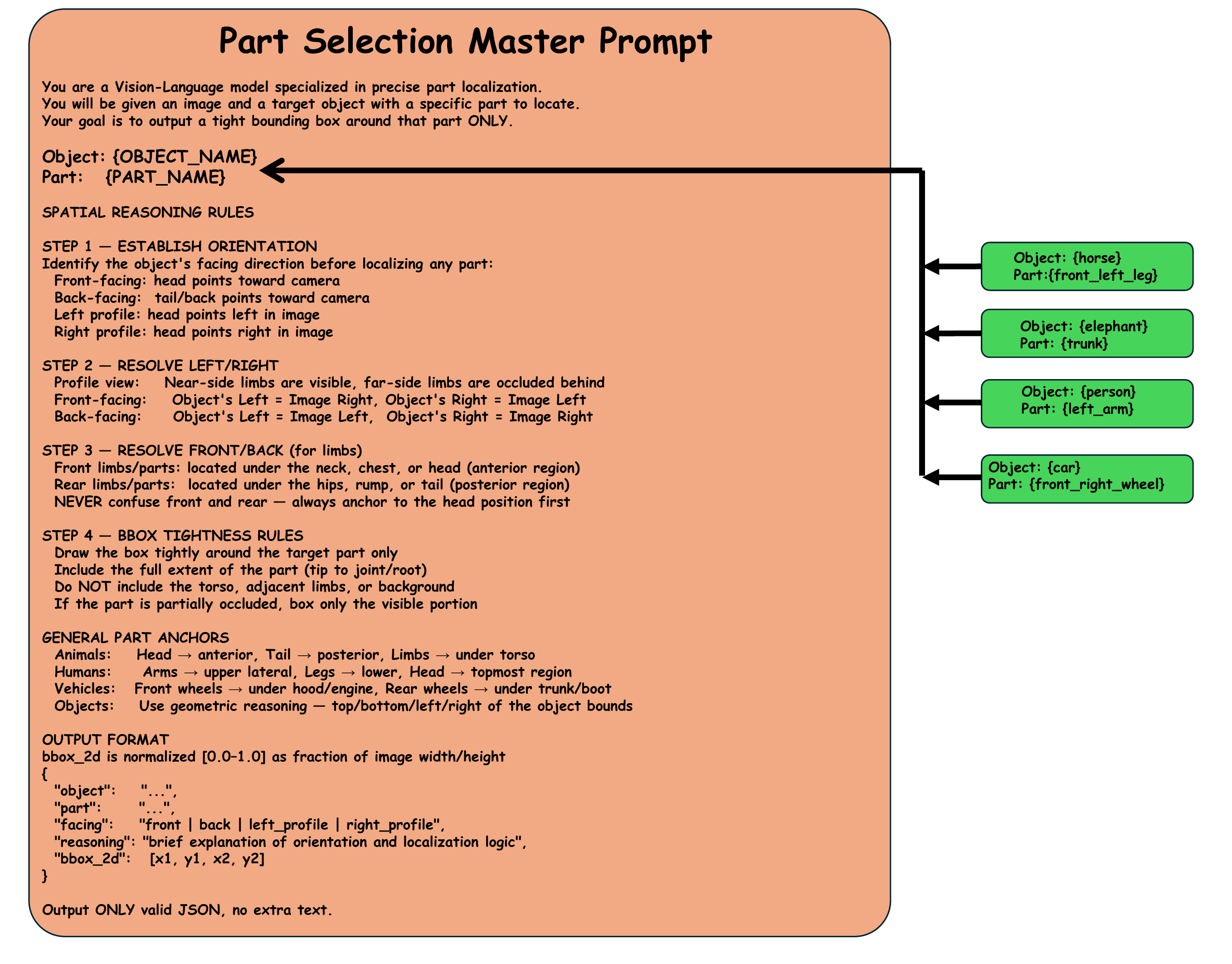}
\captionof{figure}{Part selection examples showing the mapping from
natural language part descriptions to 3D mesh regions.}
\label{fig:prompts_part_sel}
\end{strip}

\null
\twocolumn[{%
\centering
\includegraphics[width=\textwidth]{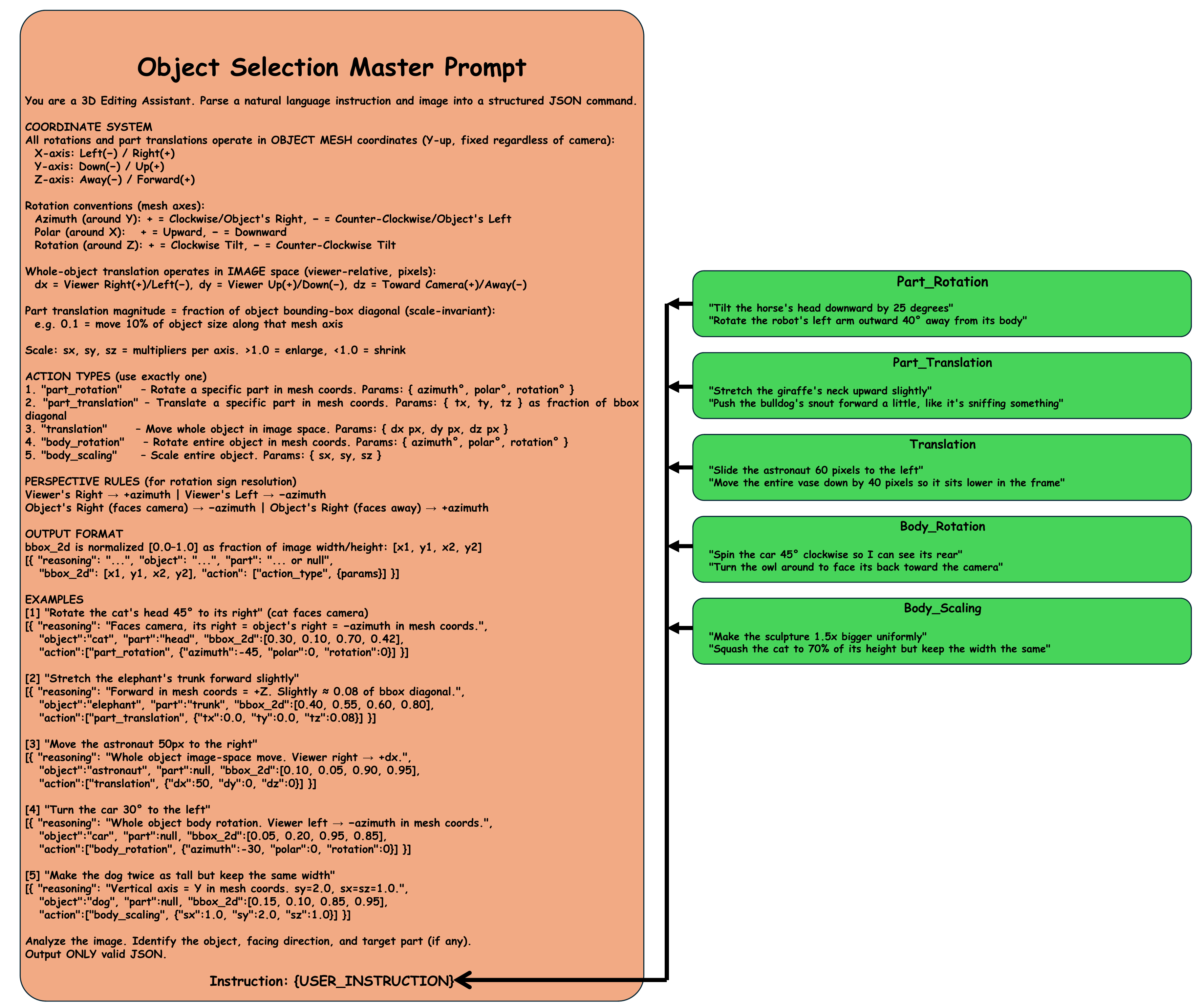}
\captionof{figure}{Object selection examples showing prompt decomposition
into localization targets across diverse object categories.}
\label{fig:prompts_obj_sel}
\vspace{1.5em}
}]

\end{document}